\PassOptionsToPackage{table}{xcolor}
\documentclass{article}
\usepackage{iclr2027_conference,times}

\usepackage[english]{babel}
\usepackage{graphicx}
\usepackage{subcaption}
\usepackage{multirow}
\usepackage{booktabs}
\usepackage{array}
\usepackage{placeins}
\usepackage{amsmath,amssymb}
\usepackage[colorlinks=true, allcolors=blue]{hyperref}
\usepackage{url}
\usepackage{utils/symbols}
\usepackage{pifont}
\def\HoffManPaperProfile{anonymous}
\def\HoffManPaperProfile{public}

\def\HoffManPublicProfileName{public}
\ifx\HoffManPaperProfile\HoffManPublicProfileName
  \InputIfFileExists{profiles/public/profile.tex}{%
    \def\HoffManActiveProfile{public}%
  }{%
    \input{profiles/anonymous}%
    \def\HoffManActiveProfile{anonymous}%
    \PackageWarningNoLine{hoffman-profile}{Public profile missing; using anonymous profile}%
  }%
\else
  \input{profiles/anonymous}%
  \def\HoffManActiveProfile{anonymous}%
\fi

\newcommand{\cmark}{\ding{51}}

\title{PredActor: Predictive Action Diffusion for Steerable Onboard Humanoid Control}

\PaperAuthorBlock

\begin{document}
\flushbottom

\maketitle
\ifx\HoffManActiveProfile\HoffManPublicProfileName
  \lhead{Preprint}
\fi

\begin{center}
\begin{minipage}{\textwidth}
  \captionsetup{type=figure}
  \centering
  \includegraphics[width=0.9\linewidth]{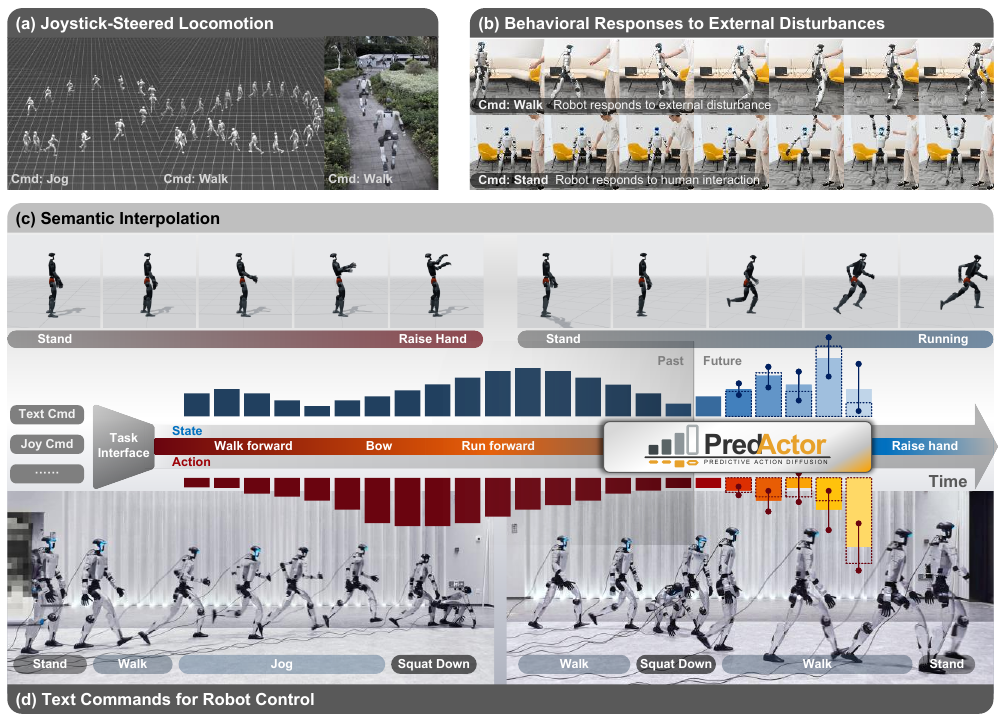}
  \caption{PredActor enables steerable and responsive humanoid control in a single directly executed policy. Unlike generator–tracker pipelines, which steer references and leave feasibility and disturbance recovery to a separate tracker, or action-only diffusion, which lacks an explicit future-state interface for test-time guidance, PredActor jointly predicts future states and executable actions. This supports (a) joystick steering, (b) disturbance-responsive behavior, (c) semantic interpolation, and (d) text-commanded control within the same closed-loop policy.}
  \label{fig:hoffman_demos}
\end{minipage}
\end{center}

\begin{abstract}
Diffusion models provide a flexible framework for motion generation, but turning this flexibility into closed-loop humanoid control remains challenging. Hierarchical generator-tracker systems steer motion through reference trajectories, yet these references may exceed the capabilities of the downstream tracker, leaving physical feasibility and disturbance recovery largely to a separate control module. Action-only diffusion avoids this separation by directly generating executable actions, but provides no explicit future-state trajectory that can be steered toward test-time motion objectives. Joint state-action diffusion offers a natural alternative, but existing controllers often rely on privileged full-body states, while learned behavior selection and test-time motion steering remain only partially integrated. We present \textbf{PredActor}, a predictive action diffusion policy that unifies these complementary forms of steerability within a single directly executed policy using only proprioceptive observations. Conditioned on proprioceptive history and optional task context, PredActor jointly predicts executable actions and an internal trajectory of future states. This predictive state trajectory provides an explicit interface for controlling the generated behavior: classifier-free guidance strengthens text-conditioned motion generation, while classifier guidance steers predicted future states toward test-time objectives. Only the predicted actions are executed, requiring neither a separate motion-reference tracker nor privileged full-body states. In simulation, PredActor reaches 44 of 45 destination targets and achieves a text retrieval score of 0.539, compared with 0.424 for conditional action diffusion, while maintaining similar disturbance survival. To make guided diffusion practical for onboard control, we further introduce rolling denoising and computation-preserving runtime optimizations, reducing the complete control callback to 16.790~ms median and 19.383~ms p95 on a Jetson Orin NX, both within the 20~ms control period. We deploy PredActor on a Unitree G1 and demonstrate, in both simulation and physical hardware, text-conditioned motion, disturbance-responsive behavior, joystick steering, and semantic interpolation within the same closed-loop policy.
\PaperProjectStatement
\end{abstract}

\section{Introduction}
\label{sec:introduction}

Humanoid policies need to follow task commands while adapting to physical feedback. Advances in motion tracking and behavior foundation models have expanded robot skills and task interfaces~\citep{luo2023perpetual,robotparty2026mimiclite,luo2025sonic,chen2026holomotion,li2025bfmzero,zeng2025bfm}. Diffusion models offer flexible conditioning for diverse motion and action generation~\citep{tevet2023human,chi2023diffusion}. Figure~\ref{fig:hoffman_demos} illustrates both requirements: joystick steering and semantic interpolation in (a,c), and disturbance response and text-command execution in (b,d). These capabilities raise a question: how can a policy guide future motion while directly generating executable actions?

\begin{center}
\begin{minipage}{\textwidth}
    \captionsetup{type=figure}
    \centering
    \includegraphics[width=\textwidth]{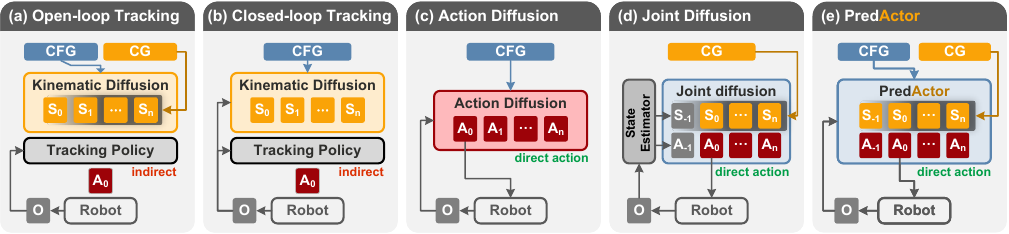}
    \caption{Paradigms of diffusion-based humanoid control. (a,b) Generator–tracker methods execute generated motion through a separate tracker. (c) Action diffusion predicts actions directly but lacks future-state guidance. (d) Joint diffusion predicts states and actions but require additional state estimation. (e) PredActor directly executes actions while using internal future-state prediction for CG and CFG-based steering. O: observation; CG/CFG: classifier/classifier-free guidance.}
    \label{fig:control_forms}
\end{minipage}
\end{center}

Diffusion control paradigms differ in how motion generation informs action execution (Figure~\ref{fig:control_forms}). Kinematic generators supply references to trackers. Open-loop variants operate \textbf{without robot feedback}~\citep{jiang2024textop,zhao2026ardy,wang2026motionbricks}, whereas closed-loop variants incorporate robot observations~\citep{tevet2024closd,chen2026reactivebfm}. Both may generate \textbf{dynamically infeasible motion outside tracker support}. Because these references are executed by a separate tracker, responses to environmental disturbances primarily reflect tracker-level recovery toward the reference trajectory rather than behavioral adaptation. \textbf{Separate planning and control clocks can also complicate synchronization and replanning}~\citep{huang2025diffusecloc,gu2026refinedp}. Action-only diffusion bypasses reference tracking by generating actions directly~\citep{chi2023diffusion,truong2024physics,huang2024diffuseloco,hoeg2024streaming}. Its predictions, however, contain no explicit future-state trajectory for classifier guidance (CG).

Joint state--action diffusion provides this predictive representation~\citep{huang2025diffusecloc,carroll2026scdp,zhang2026script}, but representative controllers often condition on privileged full-body states. That dependence makes sim-to-real transfer rely on a deployable state estimator, while the representation's complementary interfaces remain unevenly supported: classifier-free guidance (CFG) strengthens learned behavior conditions, whereas classifier guidance (CG) steers predicted motion toward test-time objectives, yet CFG is uncommon among representative joint controllers~\citep{huang2025diffusecloc,liao2025beyondmimic}. The open question is therefore functional: can a directly executed policy use deployable observations to combine learned behavior selection and test-time motion steering while retaining closed-loop disturbance robustness?

We introduce \textbf{PredActor}, a predictive action diffusion policy within this joint formulation (Figure~\ref{fig:control_forms}(e)). It jointly denoises future states and actions from proprioceptive history and optional task context. Predicted states remain internal and provide the representation for CG; conditional and null predictions support CFG. Only the selected action is issued to the joint controller, without a separate motion-reference tracker. To make guided joint diffusion practical onboard, PredActor combines rolling denoising with computation-preserving runtime optimizations to reduce iterative-inference latency.

In simulation, PredActor reaches 44 of 45 destination targets and achieves 0.539 text retrieval versus 0.424 for conditional action diffusion, with similar observed push survival (0.511 versus 0.563). On Jetson Orin NX, the complete callback has a median latency of 16.790~ms and a p95 latency of 19.383~ms, both below the 20~ms control period. \textbf{To our knowledge, PredActor is the first joint state--action diffusion policy deployed entirely on a Unitree G1's onboard Jetson Orin NX for 50~Hz control}. Figure~\ref{fig:hoffman_demos}(a,c) demonstrates simulated joystick steering and semantic interpolation, while (b,d) shows physical G1 disturbance response and text-command execution.

Our contributions are \textbf{(1) a predictive action diffusion policy for steerable humanoid control}, using only proprioceptive observations to simplify deployment and supporting both CG and CFG for task steering while retaining direct action execution; \textbf{(2) an end-to-end training and data-generation system} that turns a labeled motion library into asynchronous tracker rollouts, applies iterative data aggregation as a standard training recipe, and provides an editable task-control interface in an open-source implementation; and \textbf{(3) the first fully onboard deployment of a joint state--action diffusion policy on a Unitree G1's Jetson Orin NX for 50~Hz control}, supported by rolling inference, runtime optimization, and delay compensation.

\section{Related Work}
\label{sec:related_works}

\paragraph{Kinematic Motion Generation and Tracking.}
Physics-trained trackers execute diverse reference motions~\citep{luo2023perpetual,robotparty2026mimiclite,luo2025sonic,chen2026holomotion}, while diffusion generators provide multimodal, commandable kinematic synthesis~\citep{tevet2023human}. Related systems range from hierarchical generator--tracker stacks to unified diffusion planning and control~\citep{jiang2024textop,zhao2026ardy,wang2026motionbricks,wu2025uniphys}. Figure~\ref{fig:control_forms}(a,b) and Table~\ref{tab:capability-comparison} classify generator--tracker systems by feedback and deployment: open-loop variants do not return robot feedback to the generator; CLoSD and ReactiveBFM instead return executed state to later planning windows~\citep{tevet2024closd,chen2026reactivebfm}. Both retain interface limitations: references may violate dynamics or tracker support, and separate planning/control clocks delay synchronization and replanning; feedback mitigates drift but does not enforce dynamic feasibility during generation. Generalist BFMs broaden skills and task interfaces but retain the generation--execution split~\citep{li2025bfmzero,zeng2025bfm,shao2025langwbc}. Recent variants replace explicit kinematic references with a generated motion latent or directly generate closed-loop joint-position actions~\citep{li2026roboghost,wang2026sentinel}.

\paragraph{Diffusion Policies for Humanoid Control.}
Diffusion control also spans policies that map observations and task conditions directly to actions~\citep{chi2023diffusion,truong2024physics,huang2024diffuseloco}. Action-only diffusion, however, predicts no accompanying state trajectory for state objectives at test time. Joint state--action diffusion restores this structure and enables state-based shaping of executable actions~\citep{janner2022planning,huang2025diffusecloc,liao2025beyondmimic}. Figure~\ref{fig:control_forms}(c--e) and Table~\ref{tab:capability-comparison} organize these policies by execution paradigm, steering interfaces, and deployment. For guidance, classifier-free guidance mixes conditional and null predictions, whereas classifier guidance adds gradients from a test-time objective~\citep{ho2022classifier}. Diffuse-CLoC and BeyondMimic emphasize classifier guidance~\citep{huang2025diffusecloc,liao2025beyondmimic}; SCDP addresses partial observation but reports neither classifier guidance, classifier-free guidance, nor rolling reuse; and SCRIPT adds language conditioning and post-training but is evaluated in simulation~\citep{carroll2026scdp,zhang2026script}. Beyond modeling and conditioning, deployment requires efficient closed-loop inference. Replanning with overlapping diffusion horizons can add latency and induce cross-tick disagreement (e.g., fixed action chunks delay feedback and ensembling can average incompatible modes)~\citep{chi2023diffusion,zhao2023learning}. Fast samplers and policy distillation reduce denoising steps~\citep{2022Song_DDIM,lu2022dpm,2023Song_ConsistencyModels,prasad2024consistencypolicy}, and streaming methods reuse partially denoised horizons across ticks~\citep{hoeg2024streaming,zhang2024tedi,chen2024diffusion,chen2025falcon}, but joint state--action prediction additionally requires aligned schedules and reuse at compatible noise levels. PredActor combines proprioceptive joint state--action diffusion, both guidance interfaces, and schedule-matched rolling inference in a fully onboard policy (Table~\ref{tab:capability-comparison}). Appendix~\ref{app:extended_related_work} records the evidence.

\begin{table*}[t]
  \centering
  \caption{Capabilities and deployment of humanoid and legged generative control systems, grouped by the paradigms in Figure~\ref{fig:control_forms}. Full hardware specifications, inference timings, and source evidence appear in Table~\ref{tab:capability-deployment-full} and Appendix~\ref{app:deployment_comparison}.}
  \label{tab:capability-comparison}
  \setlength{\tabcolsep}{1.5pt}
  \renewcommand{\arraystretch}{1.12}
  \scriptsize
  \begin{tabular*}{\textwidth}{@{\extracolsep{\fill}}lclcccccll@{}}
    \toprule
    & \multicolumn{3}{c}{Model settings} & \multicolumn{3}{c}{Task capability} & \multicolumn{3}{c}{Deployment} \\
    \cmidrule(lr){2-4}\cmidrule(lr){5-7}\cmidrule(lr){8-10}
    Method & Fig.~\ref{fig:control_forms} & Paradigm & Proprio. & CFG & CG & Text & Robot & Inference GPU & Control rate \\
    \midrule
    MotionBricks + SONIC & (a) & Generator + tracker & $\diamond$ & -- & -- & -- & \cmark & Orin (on) & 50 Hz tracker$^{a}$ \\
    ARDY + SONIC & (a) & Open-loop track & $\diamond$ & $\diamond$ & -- & $\diamond$ & -- & RTX 4090$^{b}$ & -- \\
    TextOp & (a) & Open-loop track & $\diamond$ & \cmark & -- & \cmark & \cmark & RTX 4090 (off)$^{c}$ & 50 Hz tracker \\
    RoboGhost & (a) & Latent gen. + tracker & \cmark & -- & -- & \cmark & \cmark & Orin NX (part.)$^{d}$ & 50 Hz tracker \\
    ReactiveBFM & (b) & Closed-loop track & $\diamond$ & -- & -- & \cmark & \cmark & RTX 4090 (off) & 50 Hz \\
    CLoSD & (b) & Closed-loop track & -- & \cmark & -- & \cmark & -- & -- & N/A \\
    DiffuseLoco & (c) & Action diffusion & \cmark & -- & -- & -- & \cmark & RTX 4060M (on) & 30 Hz target \\
    SENTINEL & (c) & Action flow + residual & \cmark & \cmark & -- & \cmark & \cmark & RTX 4090 (off) & 50 Hz \\
    Diffuse-CLoC & (d) & Joint diffusion & -- & -- & \cmark & -- & -- & RTX 4060 (sim) & N/A \\
    BeyondMimic & (d) & Latent joint diffusion + dec. & $\diamond$ & -- & \cmark & -- & \cmark & RTX 4060M (on) & 25 Hz \\
    SCDP & (d) & Joint diffusion & \cmark & -- & -- & -- & \cmark & RTX 5090 (off) & 50 Hz \\
    SCRIPT & (d) & Joint diffusion & -- & \cmark & -- & \cmark & -- & -- & N/A \\
    UniPhys & (d) & Latent joint diff. + dec. & -- & \cmark & \cmark & \cmark & -- & A100 (sim) & N/A \\
    \midrule
    \textbf{PredActor} & (e) & Joint diffusion & \cmark & \cmark & \cmark & \cmark & \cmark & \textbf{Orin NX (on)} & \textbf{50 Hz} \\
    \bottomrule
  \end{tabular*}
  \par\vspace{2pt}
  \parbox{\textwidth}{\scriptsize
  \textbf{Symbols:} \cmark: supported; $\diamond$: partial/modular; $\times$: explicitly unsupported; --: unreported/unverified; N/A: not applicable.\par
  \textbf{Abbreviations:} Proprio.: proprioceptive input; dec.: decoder; Robot: physical deployment; on/off: onboard/offboard; sim: simulation; M: Mobile. Rates refer to robot control, not generation throughput.\par
  $^{a}$MotionBricks uses a non-diffusion generator; 50 Hz is from SONIC, not separately reported for the composition. $^{b}$ARDY generator-demo GPU; robot-stack rate unreported. $^{c}$TextOp external generator GPU; onboard tracker. $^{d}$RoboGhost establishes only the controller on Orin NX; the upstream text encoder and motion generator are not located.}
\end{table*}

\section{PredActor}
\label{sec:methods}

PredActor augments action diffusion with an internal future-state trajectory for task guidance. It is a joint state--action policy: predicted states expose future motion to steering objectives, and actions are executed directly. We first define this representation and its CG/CFG interfaces (Figure~\ref{fig:method_architecture}), then explain how teacher trajectories supervise it (Figure~\ref{fig:method_training}), and finally describe rolling inference and onboard execution (Figure~\ref{fig:method_deployment}). Detailed tensor layouts, schedule derivations, and system interfaces appear in Appendix~\ref{app:extended_methods}.

\subsection{Problem Formulation and Architecture}
\label{sec:prdp}

\noindent\textbf{Problem formulation.} At control time $t$, let $\mathbf{o}_t\in\mathbb{R}^{d_o}$ denote the proprioceptive observation available on the robot, $\mathbf{s}(t)\in\mathbb{R}^{d_s^{\mathrm{phys}}}$ its full-body motion state, and $\mathbf{a}(t)\in\mathbb{R}^{d_a}$ the action issued to the joint controller. Conditioned on an observation history $\mathbf{o}_{t-\ell+1:t}$ of length $\ell$ and optional task context $\mathbf{z}=[z_1,\ldots,z_n]$ of $n$ task tokens, PredActor predicts clean trajectories $\mathbf{s}^0\in\mathbb{R}^{h\times d_s}$ and $\mathbf{a}^0\in\mathbb{R}^{h\times d_a}$ over $h$ horizon positions in model coordinates. Superscripts denote diffusion noise levels ($0$ is clean), and subscripts $i\in\{1,\ldots,h\}$ index horizon positions. Given corrupted trajectories, the denoiser $F_\theta$ predicts:
\begin{equation}
 (\hat{\mathbf{s}}^0,\hat{\mathbf{a}}^0)
 =F_\theta(\mathbf{s}^{\mathbf{k}_s},\mathbf{a}^{\mathbf{k}_a},
              \mathbf{k}_s,\mathbf{k}_a;
              \mathbf{o}_{t-\ell+1:t},\mathbf{z}),
 \label{eq:prdp}
\end{equation}
where $\mathbf{k}_s,\mathbf{k}_a\in\{0,\ldots,T\}^{h}$ specify per-position noise levels, $T$ is the maximum diffusion level, and $\mathbf{s}^{\mathbf{k}_s},\mathbf{a}^{\mathbf{k}_a}$ are the corresponding noisy inputs. Full-body states supply training targets, while deployment requires only proprioceptive observations and optional task context. Predicted states remain internal guidance variables; only the selected action is issued to the joint controller, without a separate motion-reference tracker. Feature definitions and the physical-to-model state representation are detailed in Appendix~\ref{app:sec:prdp}.

\begin{figure}[t]
    \centering
    \includegraphics[width=\linewidth]{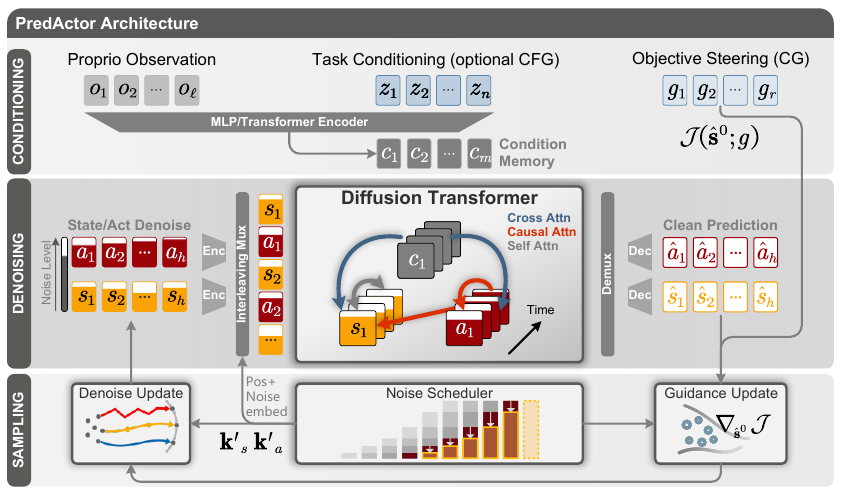}
    \caption{\textbf{Joint predictive architecture of PredActor.} Observation history and task commands condition a transformer that jointly denoises interleaved future state and action tokens. Steering changes the state update, not the current action prediction; the updated state can affect action at the next denoising step. Only actions are executed.}
    \label{fig:method_architecture}
\end{figure}

\noindent\textbf{Conditioned denoising.} Figure~\ref{fig:method_architecture} follows conditioning, denoising, and sampling from top to bottom. The encoder maps the observation-history tokens (shown as $o_1,\ldots,o_\ell$) and task tokens $z_1,\ldots,z_n$ into condition memory $\mathbf{C}=[c_1,\ldots,c_m]$, with $m$ encoded tokens. Separate projections embed noisy state and action tokens $s_i,a_i$ with their noise levels and horizon positions. We adopt Diffuse-CLoC's interleaved state--action token pairs and asymmetric self-attention pattern~\citep{huang2025diffusecloc}: state queries attend to all state tokens but not action tokens, while action queries attend causally to both streams. PredActor extends this backbone with cross-attention to $\mathbf{C}$, allowing the denoising tokens to query encoded proprioceptive history and task context. Separate heads produce clean predictions $\hat{\mathbf{s}}^0,\hat{\mathbf{a}}^0$ (shown as $\hat{s}_i,\hat{a}_i$). The noise scheduler selects per-position target levels $\mathbf{k}'_s,\mathbf{k}'_a$, and the denoising update returns both streams to the next sampling step (Section~\ref{sec:noise_schedule}).

\noindent\textbf{Guidance interfaces.} At application time, task conditioning and objective steering act as complementary interfaces. Conditional, null, and partial-condition training enables classifier-free guidance (CFG), which mixes conditional and null predictions to amplify learned task behavior (e.g., text-conditioned motion)~\citep{ho2022classifier}. Steering commands $\mathbf{g}=[g_1,\ldots,g_r]$ parameterize a differentiable objective $\mathcal{J}(\hat{\mathbf{s}}^0;\mathbf{g})$ with $r$ inputs. Classifier guidance (CG) applies $\nabla_{\hat{\mathbf{s}}^0}\mathcal{J}$ before the reverse update~\citep{huang2025diffusecloc}; it leaves the current action prediction unchanged, while the updated state can affect the next action prediction through state-to-action attention. Thus CFG reinforces learned conditions, whereas CG steers future motion toward test-time objectives. Section~\ref{sec:prdp_validation} evaluates destination steering, text control, and the predicted-state ablation (Figure~\ref{fig:architecture-comparison}; Table~\ref{tab:q1-architecture-comparison}).

\subsection{Data and Training}
\label{sec:data_workflow}
\label{sec:dagger}

\begin{figure}[t]
    \centering
    \includegraphics[width=\linewidth]{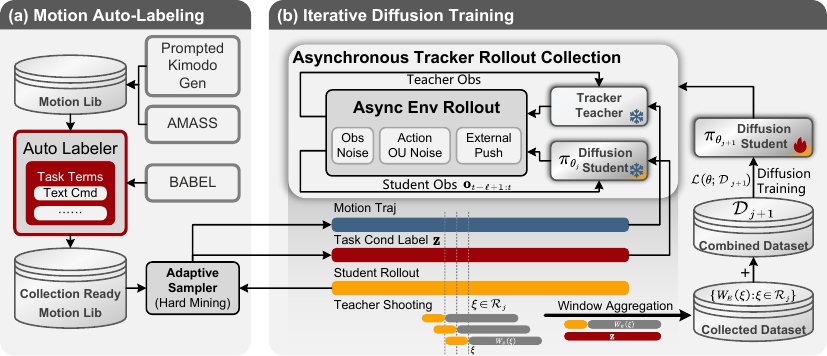}
    \caption{\textbf{Data and training pipeline.} Generated and annotated motion references drive asynchronous tracker rollouts. The resulting deployable observations and task labels condition joint state--action training. During iterative data aggregation, teacher trajectory shooting constructs target windows from learner rollout origins; the collected windows enter replay before the next reconstruction round.}
    \label{fig:method_training}
\end{figure}

Figure~\ref{fig:method_training}(a) builds the motion library from generated motion and AMASS; BABEL supplies task labels, which MotionCLIP encodes as conditioning latents~\citep{mahmood2019amass,punnakkal2021babel,tevet2022motionclip}. The adaptive sampler feeds motion references to the asynchronous rollout pipeline in (b). A frozen tracking teacher executes each reference under observation noise, action noise, and external pushes to seed $\mathcal D_0$ with observation histories, future states, teacher actions, and task labels. Training then uses a DAgger-style iterative data-aggregation recipe that alternates frozen-student rollout collection, isolated teacher trajectory shooting, and continued diffusion reconstruction~\citep{ross2011reduction}. At round $j$, the student $\pi_{\theta_j}$ induces rollout origins $\mathcal R_j$; from each origin $\xi$, an isolated teacher branch follows the aligned reference to produce a target state--action window $W_E(\xi)$, paired with the student's history and task label. These branched windows are aggregated and the student is updated for the next round:
\begin{equation}
\begin{split}
 \mathcal D_{j+1}&=\mathcal D_j\cup\{W_E(\xi):\xi\in\mathcal R_j\},\\
 \theta_{j+1}&\approx\arg\min_\theta\mathcal L(\theta;\mathcal D_{j+1}).
\end{split}
\label{eq:dagger_training}
\end{equation}
Both initial training and subsequent replay reconstruct independently corrupted state and action windows using
\begin{equation}
\begin{split}
 \mathcal{L}(\theta;\mathcal D)=\mathbb{E}\bigg[&
 \sum_{i=1}^{h} w_i^a
   \lVert\hat{\mathbf{a}}_i^0-\mathbf{a}_i^0\rVert_2^2
 +\rho\sum_{i=2}^{h}
   \lVert\Delta\hat{\mathbf{a}}_i^0-\Delta\mathbf{a}_i^0\rVert_2^2\\
 &+\gamma_s\sum_{i=1}^{h} w_i^s
   \lVert\hat{\mathbf{s}}_i^0-\mathbf{s}_i^0\rVert_2^2
 \bigg],
 \label{eq:prdp_training}
\end{split}
\end{equation}
where the expectation covers training windows, noise, and diffusion levels; $w_i^a,w_i^s$ weight horizon positions, $\Delta\mathbf{a}_i^0=\mathbf{a}_i^0-\mathbf{a}_{i-1}^0$, and $\rho,\gamma_s$ weight action-difference regularization and state reconstruction. The three terms supervise actions, action differences, and future states. The teacher supplies training targets only and is absent at deployment; dataset partitioning, branch isolation, and window alignment are detailed in Appendix~\ref{app:dagger_design}.

\subsection{Rolling Inference and Deployment}
\label{sec:noise_schedule}
\label{sec:deployment}
\label{sec:task_interface}

\begin{figure}[h]
    \centering
    \includegraphics[width=\linewidth]{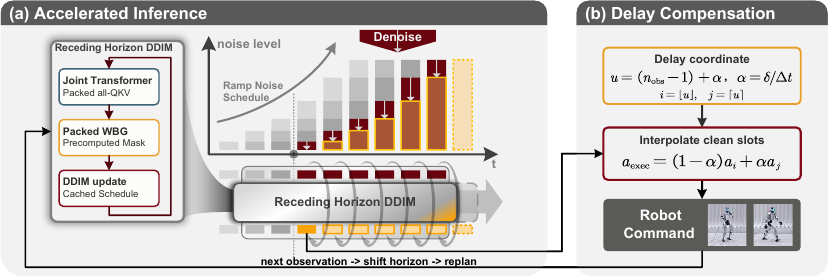}
    \caption{\textbf{Rolling inference and delay compensation.} Schedule-matched
    samples are shifted and replanned after each observation. A delay
    coordinate interpolates adjacent clean action slots before execution.}
    \label{fig:method_deployment}
\end{figure}

\noindent\textbf{Rolling denoising.} Figure~\ref{fig:method_deployment}(a) shows the receding-horizon loop that fits repeated denoising within each control period. Unlike training (independent noise-level sampling), rolling inference carries a partially denoised state--action horizon across ticks: after executing an action, buffers shift and the latest proprioceptive observation conditions the next denoise. For either stream $u\in\{s,a\}$, the DDIM update below converts a clean prediction into a reverse-sampling state at the target noise level:
\begin{equation}
 \hat\epsilon_t=\frac{u^t-\sqrt{\bar\alpha_t}\hat u^0}
 {\sqrt{1-\bar\alpha_t}},\qquad
 u^{t'}=\sqrt{\bar\alpha_{t'}}\hat u^0+
 \sqrt{1-\bar\alpha_{t'}-\sigma_t^2}\hat\epsilon_t+\sigma_t\epsilon,
 \label{eq:ddim}
\end{equation}
where $t'<t$, $\epsilon\sim\mathcal N(0,\mathbf I)$, $\bar\alpha_t$ is the cumulative noise schedule, and the DDIM variance for this jump is
\[
 \sigma_t=\eta
 \sqrt{\frac{1-\bar\alpha_{t'}}{1-\bar\alpha_t}}
 \sqrt{1-\frac{\bar\alpha_t}{\bar\alpha_{t'}}}.
\]
The evaluated deterministic profiles set $\eta=0$, so $\sigma_t=0$ and the random-noise term vanishes. The runtime can therefore carry intermediate samples across control steps instead of restarting the full horizon.

\noindent\textbf{Per-position reuse and delay compensation.} Each horizon position can take its own DDIM jump. PredActor represents the source and target levels by matrices $S,S'\in\mathbb{Z}^{K\times h}$, with each entry broadcast over the feature dimension at its horizon position; state and action streams may use different matrices. The runtime reuses the saved intermediate sample whose source level matches the next required position, adds fresh noise only to newly exposed tail positions, and denoises the shifted horizon under the latest observation. This schedule-matched reuse preserves every sample's diffusion coordinate rather than copying a clean prediction into an arbitrary noise level. In Figure~\ref{fig:method_deployment}(b), the delay coordinate $u=(n_{\mathrm{obs}}-1)+\delta/\Delta t$ selects adjacent clean action slots for interpolation, compensating actuation delay in the same horizon coordinates used for planning.

\noindent\textbf{Onboard execution.} For onboard execution, we export the diffusion actor and control interface to C++. The evaluated Orin profile uses two-step DDIM; all-QKV packing, cached condition embeddings and DDIM schedules, stacked-axis WBG, and LibTorch inference mode reduce repeated work in the callback. Section~\ref{sec:orin_acceleration} evaluates sampler speed and robustness (Figure~\ref{fig:orin-acceleration}(a,d)) and runtime optimizations (Table~\ref{tab:orin-implementation-ablation}), while Section~\ref{sec:real_robot_demos} reports physical execution. The same deployed interface supports text commands, semantic interpolation, and whole-body steering, while Appendix~\ref{app:deployment_details} specifies the export, safety, and timing boundaries.

\section{Experiments}
\label{sec:experiments}

\begin{table}[!t]
  \centering
  \caption{\textbf{Seven-policy comparison of robustness, navigation, and text control.}
  For the six learned-policy rows, entries are means $\pm$ sample SD across
  three independently trained checkpoint means; brackets show
  evaluated/requested rollout cells. ARDY + SONIC is a single-model reference.
  The CLoC+FK label denotes our evaluated Diffuse-CLoC implementation with forward-kinematics estimation of its privileged-state inputs.
  Within each numeric column, cell color runs from the worst value through a
  white midpoint to the best value, respecting the direction of the metric.}
  \label{tab:q1-architecture-comparison}
  \begingroup
  \definecolor{QOneMuted}{HTML}{6B6D69}
\begingroup
\setlength{\tabcolsep}{3.4pt}
\setlength{\arrayrulewidth}{0.50pt}
\renewcommand{\arraystretch}{1.22}
\definecolor{QOneSeparator}{HTML}{8F918C}
\def\QOneGroupRule{\hspace{2.2pt}\color{QOneSeparator}\vrule width 0.45pt\hspace{2.2pt}}
\newcommand{\QOneGroupCell}[2]{\multicolumn{#1}{!{\QOneGroupRule}c}{#2}}
\resizebox{\linewidth}{!}{%
\begin{tabular}{@{}>{\raggedright\arraybackslash}l>{\centering\arraybackslash}c>{\centering\arraybackslash}c>{\centering\arraybackslash}c>{\centering\arraybackslash}c>{\centering\arraybackslash}c>{\centering\arraybackslash}c>{\centering\arraybackslash}c>{\centering\arraybackslash}c>{\centering\arraybackslash}c>{\centering\arraybackslash}c@{}}
\hline
\multirow{2}{*}[-2.0ex]{\textbf{Method}} & \multicolumn{2}{c}{\textbf{Basic}} & \QOneGroupCell{1}{\textbf{Perturbation}} & \QOneGroupCell{4}{\textbf{Point Navigation}} & \QOneGroupCell{3}{\textbf{Text Command}} \\
\cline{2-11}
 & \shortstack{Jerk RMS\,$\downarrow$\\($10^3$ s$^{-3}$)} & \shortstack{Latency\,$\downarrow$\\(ms)} & \QOneGroupCell{1}{\shortstack{Survival\,$\uparrow$}} & \QOneGroupCell{1}{\shortstack{Survival\,$\uparrow$}} & \shortstack{Arrival\,$\uparrow$} & \shortstack{Last valid err.\,$\downarrow$\\(m)} & \shortstack{Min valid err.\,$\downarrow$\\(m)} & \QOneGroupCell{1}{\shortstack{Survival\,$\uparrow$}} & \shortstack{Retrieval\,$\uparrow$} & \shortstack{Admit.\,$\uparrow$} \\
\hline
\textbf{ARDY + SONIC} & \cellcolor[HTML]{F3C5A1}\shortstack{19.221\\{\scriptsize [20/20]}} & \cellcolor[HTML]{FFFFFF}\shortstack{\textcolor[HTML]{8A1C27}{N/E}\\{\scriptsize [0/5]}} & \QOneGroupCell{1}{\cellcolor[HTML]{F6D5BB}\shortstack{0.534\\{\scriptsize [100/100]}}} & \QOneGroupCell{1}{\cellcolor[HTML]{F2C19B}\shortstack{\textbf{1.000}\\{\scriptsize [2/15*]}}} & \cellcolor[HTML]{FFFEFD}\shortstack{0.500\\{\scriptsize [2/15*]}} & \cellcolor[HTML]{F3C39F}\shortstack{0.656\\{\scriptsize [2/15*]}} & \cellcolor[HTML]{F3C49F}\shortstack{0.656\\{\scriptsize [2/15*]}} & \QOneGroupCell{1}{\cellcolor[HTML]{F2C19B}\shortstack{\textbf{1.000}\\{\scriptsize [23/45*]}}} & \cellcolor[HTML]{C5CDD4}\shortstack{0.237\\{\scriptsize [23/45*]}} & \cellcolor[HTML]{FFFFFF}\shortstack{\textbf{0.511}\\{\scriptsize [45/45]}} \\
\textbf{DiffuseLoco w TextCFG} & \cellcolor[HTML]{F2C39E}\shortstack{15.899\\{\scriptsize $\pm$6.870 [60/60]}} & \cellcolor[HTML]{C5CDD4}\shortstack{12.141\\{\scriptsize $\pm$0.145 [15/15]}} & \QOneGroupCell{1}{\cellcolor[HTML]{F4C8A7}\shortstack{0.563\\{\scriptsize $\pm$0.009 [300/300]}}} & \QOneGroupCell{1}{\cellcolor[HTML]{FFFFFF}\shortstack{\textcolor[HTML]{8A1C27}{N/A}\\{\scriptsize [0/45]}}} & \cellcolor[HTML]{FFFFFF}\shortstack{\textcolor[HTML]{8A1C27}{N/A}\\{\scriptsize [0/45]}} & \cellcolor[HTML]{FFFFFF}\shortstack{\textcolor[HTML]{8A1C27}{N/A}\\{\scriptsize [0/45]}} & \cellcolor[HTML]{FFFFFF}\shortstack{\textcolor[HTML]{8A1C27}{N/A}\\{\scriptsize [0/45]}} & \QOneGroupCell{1}{\cellcolor[HTML]{F8DFCB}\shortstack{0.947\\{\scriptsize $\pm$0.047 [135/135]}}} & \cellcolor[HTML]{FCF0E7}\shortstack{0.424\\{\scriptsize $\pm$0.159 [126/135*]}} & \cellcolor[HTML]{FFFFFF}\shortstack{\textcolor[HTML]{8A1C27}{N/E}\\{\scriptsize [0/135]}} \\
\textbf{CLoC+FK} & \cellcolor[HTML]{C5CDD4}\shortstack{266.605\\{\scriptsize $\pm$32.412 [60/60]}} & \cellcolor[HTML]{FEF9F6}\shortstack{6.102\\{\scriptsize $\pm$0.029 [15/15]}} & \QOneGroupCell{1}{\cellcolor[HTML]{C8CFD6}\shortstack{0.303\\{\scriptsize $\pm$0.031 [300/300]}}} & \QOneGroupCell{1}{\cellcolor[HTML]{C5CDD4}\shortstack{0.064\\{\scriptsize $\pm$0.005 [45/45]}}} & \cellcolor[HTML]{C5CDD4}\shortstack{0.000\\{\scriptsize $\pm$0.000 [45/45]}} & \cellcolor[HTML]{C5CDD4}\shortstack{3.194\\{\scriptsize $\pm$0.112 [45/45]}} & \cellcolor[HTML]{C5CDD4}\shortstack{3.151\\{\scriptsize $\pm$0.053 [45/45]}} & \QOneGroupCell{1}{\cellcolor[HTML]{FFFFFF}\shortstack{\textcolor[HTML]{8A1C27}{N/A}\\{\scriptsize [0/135]}}} & \cellcolor[HTML]{FFFFFF}\shortstack{\textcolor[HTML]{8A1C27}{N/A}\\{\scriptsize [0/135]}} & \cellcolor[HTML]{FFFFFF}\shortstack{\textcolor[HTML]{8A1C27}{N/E}\\{\scriptsize [0/135]}} \\
\textbf{CLoC+FK w TextCFG} & \cellcolor[HTML]{FBEDE2}\shortstack{101.682\\{\scriptsize $\pm$2.110 [60/60]}} & \cellcolor[HTML]{E8EBEE}\shortstack{8.788\\{\scriptsize $\pm$0.141 [15/15]}} & \QOneGroupCell{1}{\cellcolor[HTML]{C5CDD4}\shortstack{0.296\\{\scriptsize $\pm$0.020 [300/300]}}} & \QOneGroupCell{1}{\cellcolor[HTML]{F5F6F7}\shortstack{0.450\\{\scriptsize $\pm$0.059 [45/45]}}} & \cellcolor[HTML]{EDEFF1}\shortstack{0.333\\{\scriptsize $\pm$0.067 [45/45]}} & \cellcolor[HTML]{F1F3F4}\shortstack{2.217\\{\scriptsize $\pm$0.450 [45/45]}} & \cellcolor[HTML]{F6F7F8}\shortstack{2.076\\{\scriptsize $\pm$0.373 [45/45]}} & \QOneGroupCell{1}{\cellcolor[HTML]{C7CFD6}\shortstack{0.782\\{\scriptsize $\pm$0.033 [135/135]}}} & \cellcolor[HTML]{FDF3EC}\shortstack{0.416\\{\scriptsize $\pm$0.106 [103/135*]}} & \cellcolor[HTML]{FFFFFF}\shortstack{\textcolor[HTML]{8A1C27}{N/E}\\{\scriptsize [0/135]}} \\
\textbf{PredActor w/o TextCFG} & \cellcolor[HTML]{F2C19B}\shortstack{\textbf{11.849}\\{\scriptsize $\pm$0.899 [60/60]}} & \cellcolor[HTML]{F3C4A0}\shortstack{1.310\\{\scriptsize $\pm$0.006 [15/15]}} & \QOneGroupCell{1}{\cellcolor[HTML]{F2C19B}\shortstack{\textbf{0.579}\\{\scriptsize $\pm$0.012 [300/300]}}} & \QOneGroupCell{1}{\cellcolor[HTML]{F6D6BC}\shortstack{0.845\\{\scriptsize $\pm$0.120 [45/45]}}} & \cellcolor[HTML]{F6D5BB}\shortstack{0.822\\{\scriptsize $\pm$0.139 [45/45]}} & \cellcolor[HTML]{F6D3B8}\shortstack{0.983\\{\scriptsize $\pm$0.301 [45/45]}} & \cellcolor[HTML]{F5D0B3}\shortstack{0.906\\{\scriptsize $\pm$0.199 [45/45]}} & \QOneGroupCell{1}{\cellcolor[HTML]{FFFFFF}\shortstack{\textcolor[HTML]{8A1C27}{N/A}\\{\scriptsize [0/135]}}} & \cellcolor[HTML]{FFFFFF}\shortstack{\textcolor[HTML]{8A1C27}{N/A}\\{\scriptsize [0/135]}} & \cellcolor[HTML]{FFFFFF}\shortstack{\textcolor[HTML]{8A1C27}{N/E}\\{\scriptsize [0/135]}} \\
\textbf{PredActor w/o state} & \cellcolor[HTML]{F2C19B}\shortstack{12.448\\{\scriptsize $\pm$0.675 [75/75]}} & \cellcolor[HTML]{F2C19B}\shortstack{\textbf{1.050}\\{\scriptsize $\pm$0.004 [15/15]}} & \QOneGroupCell{1}{\cellcolor[HTML]{FBEADC}\shortstack{0.487\\{\scriptsize $\pm$0.033 [300/300]}}} & \QOneGroupCell{1}{\cellcolor[HTML]{FFFFFF}\shortstack{\textcolor[HTML]{8A1C27}{N/A}\\{\scriptsize [0/45]}}} & \cellcolor[HTML]{FFFFFF}\shortstack{\textcolor[HTML]{8A1C27}{N/A}\\{\scriptsize [0/45]}} & \cellcolor[HTML]{FFFFFF}\shortstack{\textcolor[HTML]{8A1C27}{N/A}\\{\scriptsize [0/45]}} & \cellcolor[HTML]{FFFFFF}\shortstack{\textcolor[HTML]{8A1C27}{N/A}\\{\scriptsize [0/45]}} & \QOneGroupCell{1}{\cellcolor[HTML]{C5CDD4}\shortstack{0.778\\{\scriptsize $\pm$0.145 [135/135]}}} & \cellcolor[HTML]{FEFBF9}\shortstack{0.397\\{\scriptsize $\pm$0.023 [98/135*]}} & \cellcolor[HTML]{FFFFFF}\shortstack{\textcolor[HTML]{8A1C27}{N/E}\\{\scriptsize [0/135]}} \\
\hline
\textbf{PredActor} & \cellcolor[HTML]{F3C39F}\shortstack{16.810\\{\scriptsize $\pm$1.409 [60/60]}} & \cellcolor[HTML]{F3C4A0}\shortstack{1.321\\{\scriptsize $\pm$0.011 [15/15]}} & \QOneGroupCell{1}{\cellcolor[HTML]{F8DFCB}\shortstack{0.511\\{\scriptsize $\pm$0.030 [300/300]}}} & \QOneGroupCell{1}{\cellcolor[HTML]{F2C39F}\shortstack{0.983\\{\scriptsize $\pm$0.030 [45/45]}}} & \cellcolor[HTML]{F2C19B}\shortstack{\textbf{0.978}\\{\scriptsize $\pm$0.038 [45/45]}} & \cellcolor[HTML]{F2C19B}\shortstack{\textbf{0.605}\\{\scriptsize $\pm$0.029 [45/45]}} & \cellcolor[HTML]{F2C19B}\shortstack{\textbf{0.604}\\{\scriptsize $\pm$0.028 [45/45]}} & \QOneGroupCell{1}{\cellcolor[HTML]{FFFFFF}\shortstack{0.888\\{\scriptsize $\pm$0.068 [135/135]}}} & \cellcolor[HTML]{F2C19B}\shortstack{\textbf{0.539}\\{\scriptsize $\pm$0.075 [119/135*]}} & \cellcolor[HTML]{FFFFFF}\shortstack{\textcolor[HTML]{8A1C27}{N/E}\\{\scriptsize [0/135]}} \\
\hline
\end{tabular}%
}
\endgroup

  \vspace{0.35em}
  \begin{minipage}{\textwidth}
  \fontsize{6.4}{7.2}\selectfont\color{QOneMuted}
  The six learned policies use training seeds 92025/92026/92027. Bold marks the numeric optimum at the displayed precision, including conditional cells and ties. N/A marks an unavailable task interface; N/E marks a metric not evaluated under the common protocol, including reference admission where no admission test was performed. Neither is zero-imputed. An asterisk marks a conditional subset excluded from full-coverage comparisons. The ARDY + SONIC destination entries use the strict raw-admission subset (2/15); the separate limit-projected arm is not substituted. Latency is synchronized, post-warmup RTX~4090 actor time amortized per action at $B=5$; text retrieval uses complete windows beginning at least 3~s after command activation.
  \end{minipage}
  \endgroup
  \vspace{-0.65\baselineskip}
\end{table}

We ask three questions: \textbf{Q1}, can PredActor combine CG destination steering and CFG text control while retaining disturbance recovery? \textbf{Q2}, can exact guided inference meet the 20~ms onboard budget? \textbf{Q3}, can the same policy support simulated interfaces and physical G1 execution? The policy controls 29 actuated joints at 50~Hz from proprioceptive history and predicts a 20-step state--action horizon. Full protocols, metrics, coverage, and uncertainty treatment are in Appendix~\ref{app:experiment_details}.

\subsection{Steering and Closed-Loop Control}
\label{sec:prdp_validation}
\label{sec:baselines}

\noindent\textbf{Comparison setup.} We compare seven policies in one calibrated MuJoCo plant under matched push, destination, and text protocols (Table~\ref{tab:q1-architecture-comparison}). The six learned policies use 7,600 teacher-rollout episodes and three training seeds; ARDY+SONIC is a single-model reference. We report checkpoint means and sample SDs for survival, arrival within 0.6~m, navigation error, MotionCLIP retrieval, jerk, and latency. We denote our evaluated Diffuse-CLoC implementation as CLoC+FK; forward kinematics estimates its required privileged-state inputs, and the suffix denotes this observation interface. Appendix~\ref{app:architecture_protocol} details the baselines, training settings, and coverage.

\begin{figure}[t]
  \vspace{-\baselineskip}
  \centering
  \includegraphics[width=\textwidth]{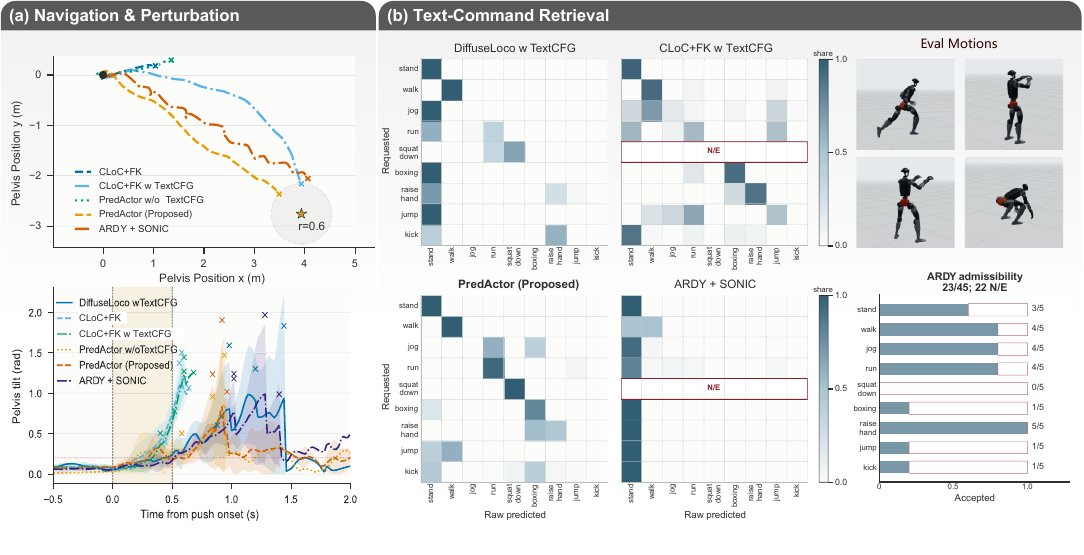}
  \caption{\textbf{Representative MuJoCo traces for six of the seven policy rows.}
  \textbf{(a)} Representative destination trajectories in the common
  protocol frame (top; the star is the target and the circle marks the
  0.6~m arrival threshold) and direction-matched pelvis tilt under a 300~N push
  (bottom; lines are means, bands show sample SD, and crosses mark early
  terminals) across 20 paired direction/seed cells per policy. \textbf{(b)}
  Per-class raw MotionCLIP top-1 retrieval for each evaluable policy, example
  evaluation motions, and ARDY+SONIC reference admissibility by requested
  class. Missing cells remain distinct from zero; 22/45 ARDY references are
  rejected before SONIC rollout. The seven-row aggregate and its
  training-replica denominators are reported in
  Table~\ref{tab:q1-architecture-comparison}.}
  \label{fig:architecture-comparison}
\end{figure}

\noindent\textbf{Steering performance.} PredActor combines destination guidance and text control (Figure~\ref{fig:architecture-comparison}(a,b)). It reaches 44/45 targets with $0.605\pm0.029$~m last-valid error and achieves $0.539\pm0.075$ text retrieval over 119/135 eligible windows, versus $0.424\pm0.159$ over 126/135 for DiffuseLoco w TextCFG. Their text survival is $0.888\pm0.068$ and $0.947\pm0.047$, respectively, over 135 trials; retrieval uses each policy's eligible windows.

\noindent\textbf{Efficiency and robustness.} PredActor combines low latency with competitive smoothness and push survival (Table~\ref{tab:q1-architecture-comparison}; Figure~\ref{fig:architecture-comparison}(a)). Actor latency is $1.321\pm0.011$~ms, versus $12.141\pm0.145$ for DiffuseLoco w TextCFG and $8.788\pm0.141$ for CLoC+FK w TextCFG. Corresponding jerk RMS values are $16.810\pm1.409$, $15.899\pm6.870$, and $101.682\pm2.110$. Push survival is $0.511\pm0.030$ versus $0.563\pm0.009$ for DiffuseLoco w TextCFG.

\noindent\textbf{State ablation.} Predicted states provide the destination-CG interface and support closed-loop behavior. Without this stream, push survival, text survival, and retrieval are $0.487\pm0.033$, $0.778\pm0.145$, and $0.397\pm0.023$ over 98/135 eligible retrieval windows, respectively, all below the full-model means (Table~\ref{tab:q1-architecture-comparison}). Together, PredActor supports both steering interfaces with low latency, competitive smoothness, and survival comparable to the baselines.

\vspace{-0.5\baselineskip}
\subsection{Inference Acceleration and Onboard Evaluation}
\label{sec:orin_acceleration}
\label{sec:stats_deployment}

\noindent\textbf{Deployment setup.} We benchmark batch-one FP32 guided inference and the complete callback on one Jetson Orin NX, varying sampler steps and implementation settings (Appendix~\ref{app:orin_details}). Timing uses received robot-state messages without issuing actuation commands. Separate Unitree G1 trials evaluate four motion transitions and two push settings, with five trials per setting; success requires completing the transition or remaining upright after the push.

\begin{table}[t]
\centering
\small
\setlength{\tabcolsep}{3pt}
\renewcommand{\arraystretch}{1.0}

\begin{minipage}[c]{0.31\linewidth}
\centering
\caption{\textbf{Real-robot trial outcomes} (five trials per setting).}
\label{tab:real-robot-quant}
\begin{tabular*}{\linewidth}{@{}l@{\extracolsep{\fill}}r@{}}
\toprule
& Success \\
\midrule
Stand$\rightarrow$Walk & 5/5 \\
Walk$\rightarrow$Jog & 5/5 \\
Stand$\rightarrow$Squat & 5/5 \\
Squat$\rightarrow$Stand & 4/5 \\
Stand push & 4/5 \\
Walk push & 4/5 \\
\bottomrule
\end{tabular*}
\par\vspace{3pt}
\begin{minipage}{\linewidth}\footnotesize\raggedright
Success: complete the requested transition or remain upright after the push. Falls, incomplete behavior, and safety stops count as failures. Five trials per setting; counts are descriptive. Per-trial sequences appear in Appendix~\ref{app:real_robot_trial_sequences}.
\end{minipage}
\end{minipage}\hfill
\begin{minipage}[c]{0.67\linewidth}
\centering
\caption[\textbf{Cumulative acceleration on Orin.}]{\textbf{Cumulative acceleration on Orin.} Frozen actor CUDA timing except the final live callback; NS-DDIM2, B=1 FP32, one forward/step (two total), WBG on, eager finite checks.}
\label{tab:orin-implementation-ablation}
\setlength{\tabcolsep}{2.5pt}
\begin{tabular*}{\linewidth}{@{}l@{\extracolsep{\fill}\hspace{3pt}}rrrr@{}}
\toprule
Composition / boundary & p50 & p95 & Saved & $>20$ ms \\
\midrule
Structural baseline; seq. WBG & 21.461 & 22.036 & -- & 600/600 \\
\quad + stacked WBG & 18.315 & 19.677 & 3.147 & 8/600 \\
\quad + DDIM precomputation & 17.442 & 18.347 & 0.872 & 0/600 \\
\quad + all-QKV packing & 16.331 & 17.372 & 1.111 & 0/600 \\
\quad + condition cache & 15.930 & 16.728 & 0.401 & 0/600 \\
\midrule
Same structure, runtime remeasurement & 16.234 & 16.912 & -- & 1/600 \\
\textbf{\quad + inference mode (final actor)} & \textbf{13.175} & \textbf{13.400} & \textbf{3.058} & \textbf{0/600} \\
\midrule
\textbf{Final live callback-to-host} & \textbf{16.790} & \textbf{19.383} & \textbf{--} & \textbf{7/600} \\
\bottomrule
\end{tabular*}
\par\vspace{3pt}
\begin{minipage}{\linewidth}\footnotesize
Times in ms: medians of three block quantiles (200 calls/block). Saved: preceding-row p50 decrease within one suite, not additive factor effects. Rules mark suite/boundary changes; Appendix~\ref{app:orin_details} distinguishes the original-export $P_0$ from the matched structural baseline and reports effect ranges. The last column counts calls exceeding the 20~ms budget.
\end{minipage}
\end{minipage}
\end{table}

\begin{figure}[!t]
  \centering
  \includegraphics[width=\textwidth]{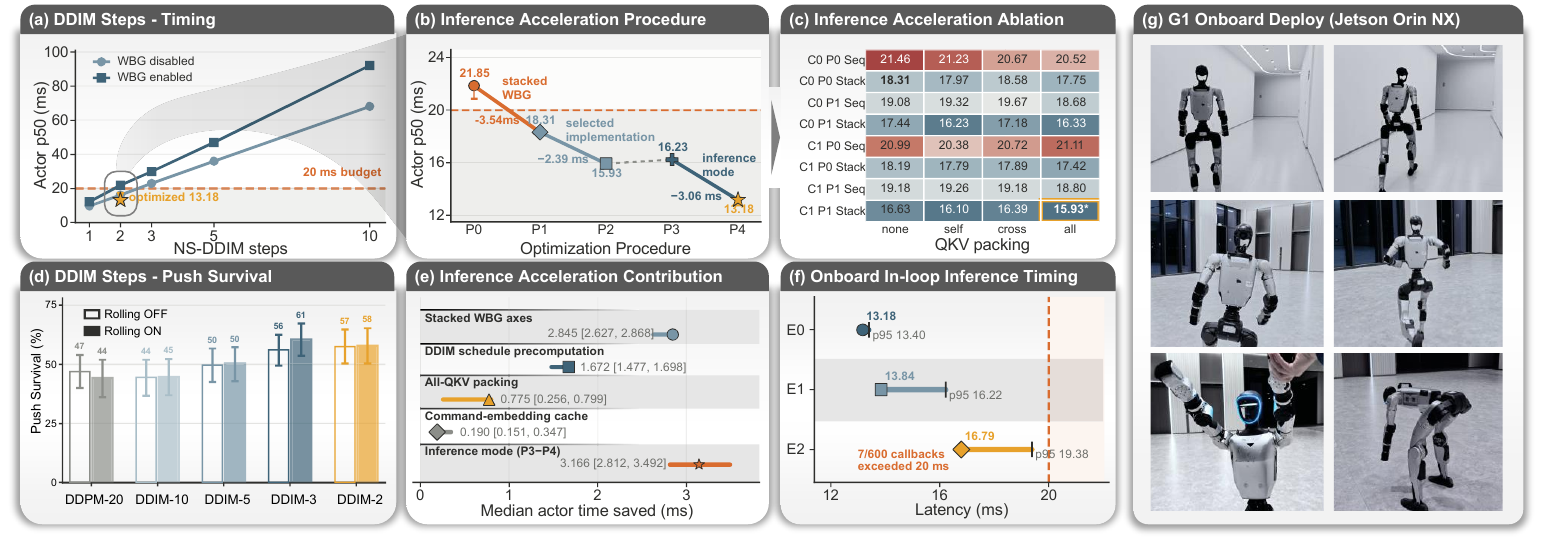}
  \caption{\textbf{Orin acceleration and onboard G1 deployment.}
  \textbf{(a,d)} Sampler timing and push survival.
  \textbf{(b)} Path $\mathrm{P}_0$--$\mathrm{P}_4$ from the original actor through cache-capable re-export, the selected configuration, runtime remeasurement, and LibTorch inference mode.
  \textbf{(c)} Ablation matrix (C: condition cache; P: schedule precomputation; Seq/Stack: WBG axes).
  \textbf{(e)} Implementation savings; markers show the median reduction across three paired blocks, and brackets give the corresponding min--max range.
  \textbf{(f)} p50--p95 for $\mathrm{E}_0$ (fixed-input actor), $\mathrm{E}_1$ (new-state receipt/conversion plus actor), and $\mathrm{E}_2$ (complete callback to host action). Each endpoint uses $3\times200$ retained calls; 593/600 $\mathrm{E}_2$ calls finish within 20~ms and p95 is 19.383~ms.
  \textbf{(g)} Qualitative frames from onboard G1 deployment with Jetson Orin NX.}
  \label{fig:orin-acceleration}
\end{figure}

\noindent\textbf{Sampling and stability.} Two-step NS-DDIM balances inference speed and disturbance recovery. Reducing ten steps to two lowers guided actor p50 from 92.011 to 21.846~ms before optimization (Figure~\ref{fig:orin-acceleration}(a)). With rolling denoising, push survival is 58\% at two steps, close to 61\% at three and above 45\% at ten (Figure~\ref{fig:orin-acceleration}(d)), supporting the two-step deployment choice.

\noindent\textbf{Acceleration contributions.} Complementary optimizations reduce actor p50 to 13.175~ms (Figure~\ref{fig:orin-acceleration}(b,c)). Matched ablations attribute median savings of 2.845~ms to WBG stacking, 1.672~ms to schedule precomputation, 0.775~ms to all-QKV packing, and 0.190~ms to condition caching; inference mode saves 3.166~ms across paired runtime blocks (Figure~\ref{fig:orin-acceleration}(e)). These configuration-dependent effects are not additive; Table~\ref{tab:orin-implementation-ablation} reports the cumulative path. Output drift remains below $1.073\times10^{-6}$.

\noindent\textbf{Onboard evaluation.} The complete callback meets the 20~ms budget at p95: p50/p95 are 16.790/19.383~ms, with 593/600 calls within budget (Figure~\ref{fig:orin-acceleration}(f)). Physical G1 trials in Table~\ref{tab:real-robot-quant} achieve 5/5 successes for stand-to-walk, walk-to-jog, and stand-to-squat, and 4/5 for squat-to-stand and both push settings. Together, these measurements support 50~Hz onboard control at the measured p95 and repeatable physical transitions and recovery.

\subsection{Onboard and Simulation Demonstrations}
\label{sec:real_robot_demos}

\noindent\textbf{Text control.} Figure~\ref{fig:hoffman_demos}(d) shows staged stand, walk, jog, and squat commands on the physical G1.
\textbf{Joystick control.} Figure~\ref{fig:hoffman_demos}(a) illustrates joystick-steered locomotion.
\textbf{Semantic interpolation.} Figure~\ref{fig:hoffman_demos}(c) shows simulated interpolation from standing toward raising a hand or running.
\textbf{Interactive responses.} Figure~\ref{fig:hoffman_demos}(b) shows selected physical responses to external disturbances, while Figure~\ref{fig:behavior-response-appendix} adds four external-interaction sequences under active walk or stand commands. Table~\ref{tab:real-robot-quant} separately reports five trials for four transitions and two push settings, and Appendix~\ref{app:real_robot_trial_sequences} shows every outcome under the same scoring rule. Figure~\ref{fig:orin-acceleration}(g) provides qualitative frames confirming execution on the G1--Jetson Orin NX stack; Appendix~\ref{app:deployment_demo_protocol} specifies the demonstration scope.

\FloatBarrier

\section{Conclusion}
\label{sec:conclusion}

PredActor unifies predictive motion generation and direct action execution in a joint state--action diffusion policy for steerable humanoid control. Predicted future states form a differentiable interface for classifier guidance, classifier-free guidance selects task-conditioned behaviors, and paired actions close the loop from proprioceptive history. This eliminates a separate motion tracker while retaining test-time steering.
Across matched simulations, PredActor reaches 44 of 45 destination targets, achieves 0.539 text retrieval, and maintains competitive disturbance survival and smoothness. Runtime optimization yields 16.790~ms median and 19.383~ms p95 complete-callback latency on Jetson Orin NX, while onboard G1 execution demonstrates the deployable interface across six physical settings.
These results show that internal predictive trajectories can connect high-level steering with responsive low-level execution. Future work will strengthen guidance, expand language--motion coverage, and support richer multi-contact behaviors.

\clearpage
\raggedbottom
\subsection*{Reproducibility statement}
The anonymous project page linked in the abstract provides the open-source code
and reproduction materials for PredActor. Appendix~\ref{app:extended_methods}
documents the model, training, and deployment details;
Appendix~\ref{app:experiment_details} specifies the evaluation protocols,
metrics, trial coverage, timing boundaries, and failure handling; and
Appendix~\ref{app:complementary_results} reports complementary ablations and
limitations.

\subsection*{AI use statement}
Generative AI tools assisted with conceptual framing, experiment-design review,
literature discovery and summarization, manuscript editing, and code-based
figures and scripts. We verified literature claims against primary sources,
inspected generated code and artifacts, rebuilt the manuscript, and reviewed all
AI-assisted content. Generative AI output was not used as experimental evidence.
The authors take responsibility for the final text, claims, and artifacts.

\subsection*{Ethics statement}
This work does not collect or analyze human-subject data. Real-robot
experiments were conducted by trained operators using appropriate safety
procedures, and identifiable faces in released visual materials are blurred.

\bibliographystyle{iclr2027_conference}
\bibliography{library}

@inproceedings{tevet2022motionclip,
  title        = {{MotionCLIP}: Exposing human motion generation to {CLIP} space},
  author       = {Tevet, Guy and Gordon, Brian and Hertz, Amir and Bermano, Amit H and Cohen-Or, Daniel},
  booktitle    = {European Conference on Computer Vision (ECCV)},
  pages        = {358--374},
  publisher    = {Springer},
  year         = {2022},
  doi          = {10.1007/978-3-031-20047-2_21},
  url          = {https://doi.org/10.1007/978-3-031-20047-2_21},
  organization = {Springer}
}

@inproceedings{mahmood2019amass,
  title     = {{AMASS}: Archive of Motion Capture as Surface Shapes},
  author    = {Mahmood, Naureen and Ghorbani, Nima and Troje, Nikolaus F. and Pons-Moll, Gerard and Black, Michael J.},
  booktitle = {Proceedings of the IEEE/CVF International Conference on Computer Vision (ICCV)},
  pages     = {5442--5451},
  year      = {2019},
  doi       = {10.1109/ICCV.2019.00554},
  url       = {https://doi.org/10.1109/ICCV.2019.00554}
}

@inproceedings{punnakkal2021babel,
  title     = {{BABEL}: Bodies, Action and Behavior with English Labels},
  author    = {Punnakkal, Abhinanda R. and Chandrasekaran, Arjun and Athanasiou, Nikos and Quiros-Ramirez, Alejandra and Black, Michael J.},
  booktitle = {Proceedings of the IEEE/CVF Conference on Computer Vision and Pattern Recognition (CVPR)},
  pages     = {722--731},
  year      = {2021},
  doi       = {10.1109/CVPR46437.2021.00078},
  url       = {https://doi.org/10.1109/CVPR46437.2021.00078}
}

@inproceedings{luo2023perpetual,
  title     = {Perpetual Humanoid Control for Real-time Simulated Avatars},
  author    = {Luo, Zhengyi and Cao, Jinkun and Winkler, Alexander and Kitani, Kris and Xu, Weipeng},
  booktitle = {2023 IEEE/CVF International Conference on Computer Vision (ICCV)},
  pages     = {10895--10904},
  publisher = {IEEE},
  year      = {2023},
  month     = {Oct},
  doi       = {10.1109/iccv51070.2023.01000},
  url       = {https://doi.org/10.1109/iccv51070.2023.01000}
}

@inproceedings{he2024h2o,
  title     = {Learning Human-to-Humanoid Real-Time Whole-Body Teleoperation},
  author    = {He, Tairan and Luo, Zhengyi and Xiao, Wenli and Zhang, Chong and Kitani, Kris and Liu, Changliu and Shi, Guanya},
  booktitle = {2024 IEEE/RSJ International Conference on Intelligent Robots and Systems (IROS)},
  pages     = {8944--8951},
  publisher = {IEEE},
  year      = {2024},
  month     = {Oct},
  doi       = {10.1109/iros58592.2024.10801984},
  url       = {https://doi.org/10.1109/iros58592.2024.10801984}
}

@inproceedings{he2024omni,
  title     = {{OmniH2O}: Universal and Dexterous Human-to-Humanoid Whole-Body Teleoperation and Learning},
  author    = {He, Tairan and Luo, Zhengyi and He, Xialin and Xiao, Wenli and Zhang, Chong and Zhang, Weinan and Kitani, Kris M. and Liu, Changliu and Shi, Guanya},
  booktitle = {Proceedings of the 8th Conference on Robot Learning},
  series    = {Proceedings of Machine Learning Research},
  volume    = {270},
  pages     = {1516--1540},
  publisher = {PMLR},
  year      = {2025},
  month     = {06--09 Nov},
  url       = {https://proceedings.mlr.press/v270/he25b.html}
}

@inproceedings{ze2024twist,
  title     = {{TWIST}: Teleoperated Whole-Body Imitation System},
  author    = {Ze, Yanjie and Chen, Zixuan and Ara{\'u}jo, Jo{\~a}o Pedro and Cao, Zi-ang and Peng, Xue Bin and Wu, Jiajun and Liu, C. Karen},
  booktitle = {Proceedings of The 9th Conference on Robot Learning},
  series    = {Proceedings of Machine Learning Research},
  volume    = {305},
  pages     = {2143--2154},
  publisher = {PMLR},
  year      = {2025},
  month     = {27--30 Sep},
  url       = {https://proceedings.mlr.press/v305/ze25a.html}
}

@inproceedings{ho2022classifier,
  title         = {Classifier-Free Diffusion Guidance},
  author        = {Jonathan Ho and Tim Salimans},
  booktitle     = {NeurIPS 2021 Workshop on Deep Generative Models and Downstream Applications},
  year          = {2021},
  url           = {https://openreview.net/forum?id=qw8AKxfYbI}
}

@inproceedings{chi2023diffusion,
  title     = {Diffusion Policy: Visuomotor Policy Learning via Action Diffusion},
  author    = {Chi, Cheng and Feng, Siyuan and Du, Yilun and Xu, Zhenjia and Cousineau, Eric and Burchfiel, Benjamin C. M. and Song, Shuran},
  booktitle = {Robotics: Science and Systems XIX},
  series    = {RSS2023},
  publisher = {Robotics: Science and Systems Foundation},
  year      = {2023},
  month     = {July},
  doi       = {10.15607/rss.2023.xix.026},
  url       = {https://doi.org/10.15607/rss.2023.xix.026}
}

@inproceedings{janner2022planning,
  title     = {Planning with Diffusion for Flexible Behavior Synthesis},
  author    = {Janner, Michael and Du, Yilun and Tenenbaum, Joshua B. and Levine, Sergey},
  booktitle = {Proceedings of the 39th International Conference on Machine Learning},
  series    = {Proceedings of Machine Learning Research},
  volume    = {162},
  pages     = {9902--9915},
  publisher = {PMLR},
  year      = {2022},
  month     = {17--23 Jul},
  url       = {https://proceedings.mlr.press/v162/janner22a.html}
}

@inproceedings{ajay2023conditional,
  title     = {Is Conditional Generative Modeling all you need for Decision Making?},
  author    = {Anurag Ajay and Yilun Du and Abhi Gupta and Joshua B. Tenenbaum and Tommi S. Jaakkola and Pulkit Agrawal},
  booktitle = {International Conference on Learning Representations},
  year      = {2023},
  url       = {https://openreview.net/forum?id=sP1fo2K9DFG}
}

@inproceedings{huang2024diffuseloco,
  title     = {{DiffuseLoco}: Real-Time Legged Locomotion Control with Diffusion from Offline Datasets},
  author    = {Huang, Xiaoyu and Chi, Yufeng and Wang, Ruofeng and Li, Zhongyu and Peng, Xue Bin and Shao, Sophia and Nikolic, Borivoje and Sreenath, Koushil},
  booktitle = {Proceedings of the 8th Conference on Robot Learning},
  series    = {Proceedings of Machine Learning Research},
  volume    = {270},
  pages     = {1567--1589},
  publisher = {PMLR},
  year      = {2025},
  month     = {06--09 Nov},
  url       = {https://proceedings.mlr.press/v270/huang25a.html}
}

@inproceedings{truong2024physics,
  title     = {{PDP}: Physics-Based Character Animation via Diffusion Policy},
  author    = {Truong, Takara Everest and Piseno, Michael and Xie, Zhaoming and Liu, C. Karen},
  booktitle = {SIGGRAPH Asia 2024 Conference Papers},
  series    = {SA '24},
  pages     = {1--10},
  publisher = {ACM},
  year      = {2024},
  month     = {Dec},
  doi       = {10.1145/3680528.3687683},
  url       = {https://doi.org/10.1145/3680528.3687683},
  articleno = {86},
  numpages  = {10}
}

@inproceedings{tevet2024closd,
  title     = {{CLoSD}: Closing the Loop between Simulation and Diffusion for multi-task character control},
  author    = {Tevet, Guy and Raab, Sigal and Cohan, Setareh and Reda, Daniele and Luo, Zhengyi and Peng, Xue Bin and Bermano, Amit H. and van de Panne, Michiel},
  booktitle = {International Conference on Learning Representations},
  volume    = {2025},
  pages     = {46506--46520},
  year      = {2025},
  url       = {https://openreview.net/forum?id=pZISppZSTv}
}

@inproceedings{tevet2023human,
  title     = {Human Motion Diffusion Model},
  author    = {Tevet, Guy and Raab, Sigal and Gordon, Brian and Shafir, Yonatan and Cohen-Or, Daniel and Bermano, Amit H.},
  booktitle = {International Conference on Learning Representations},
  year      = {2023},
  url       = {https://openreview.net/forum?id=SJ1kSyO2jwu}
}

@inproceedings{zhao2023learning,
  title     = {Learning Fine-Grained Bimanual Manipulation with Low-Cost Hardware},
  author    = {Zhao, Tony Z. and Kumar, Vikash and Levine, Sergey and Finn, Chelsea},
  booktitle = {Robotics: Science and Systems XIX},
  series    = {RSS2023},
  publisher = {Robotics: Science and Systems Foundation},
  year      = {2023},
  month     = {July},
  doi       = {10.15607/rss.2023.xix.016},
  url       = {https://doi.org/10.15607/rss.2023.xix.016}
}

@inproceedings{zhang2024tedi,
  title     = {{TEDi}: Temporally-Entangled Diffusion for Long-Term Motion Synthesis},
  author    = {Zhang, Zihan and Liu, Richard and Aberman, Kfir and Hanocka, Rana},
  booktitle = {Special Interest Group on Computer Graphics and Interactive Techniques Conference Conference Papers},
  series    = {SIGGRAPH '24},
  pages     = {1--11},
  publisher = {ACM},
  year      = {2024},
  month     = {July},
  doi       = {10.1145/3641519.3657515},
  url       = {https://doi.org/10.1145/3641519.3657515}
}

@inproceedings{hoeg2024streaming,
  title     = {Fast Policy Synthesis with Variable Noise Diffusion Models},
  author    = {H{\o}eg, Sigmund H. and Du, Yilun and Egeland, Olav},
  booktitle = {2025 IEEE International Conference on Robotics and Automation (ICRA)},
  pages     = {4821--4828},
  publisher = {IEEE},
  year      = {2025},
  doi       = {10.1109/ICRA55743.2025.11127858},
  url       = {https://doi.org/10.1109/ICRA55743.2025.11127858}
}

@inproceedings{chen2024diffusion,
  title     = {Diffusion Forcing: Next-token Prediction Meets Full-Sequence Diffusion},
  author    = {Chen, Boyuan and Mons{\'o}, Diego Mart{\'i} and Du, Yilun and Simchowitz, Max and Tedrake, Russ and Sitzmann, Vincent},
  booktitle = {Advances in Neural Information Processing Systems 37},
  series    = {NeurIPS 2024},
  pages     = {24081--24125},
  publisher = {Neural Information Processing Systems Foundation, Inc. (NeurIPS)},
  year      = {2024},
  doi       = {10.52202/079017-0759},
  url       = {https://doi.org/10.52202/079017-0759}
}

@inproceedings{wang2025onedp,
  title     = {One-Step Diffusion Policy: Fast Visuomotor Policies via Diffusion Distillation},
  author    = {Wang, Zhendong and Li, Max and Mandlekar, Ajay and Xu, Zhenjia and Fan, Jiaojiao and Narang, Yashraj and Fan, Linxi and Zhu, Yuke and Balaji, Yogesh and Zhou, Mingyuan and Liu, Ming-Yu and Zeng, Yu},
  booktitle = {Proceedings of the 42nd International Conference on Machine Learning},
  series    = {Proceedings of Machine Learning Research},
  volume    = {267},
  pages     = {63399--63416},
  publisher = {PMLR},
  year      = {2025},
  month     = {13--19 Jul},
  url       = {https://proceedings.mlr.press/v267/wang25ba.html}
}

@inproceedings{romer2025dpcc,
  title     = {Diffusion Predictive Control with Constraints},
  author    = {R{\"o}mer, Ralf and Rohr, Alexander von and Schoellig, Angela},
  booktitle = {Proceedings of the 7th Annual Learning for Dynamics \& Control Conference},
  series    = {Proceedings of Machine Learning Research},
  volume    = {283},
  pages     = {791--803},
  publisher = {PMLR},
  year      = {2025},
  month     = {04--06 Jun},
  url       = {https://proceedings.mlr.press/v283/romer25a.html}
}

@article{liao2025beyondmimic,
  title     = {{BeyondMimic}: From motion tracking to versatile humanoid control via guided diffusion},
  author    = {Liao, Qiayuan and Truong, Takara E. and Huang, Xiaoyu and Gao, Yuman and Tevet, Guy and Sreenath, Koushil and Liu, C. Karen},
  journal   = {Science Robotics},
  volume    = {11},
  number    = {117},
  publisher = {American Association for the Advancement of Science (AAAS)},
  year      = {2026},
  month     = {Aug},
  doi       = {10.1126/scirobotics.adx8924},
  url       = {https://doi.org/10.1126/scirobotics.adx8924}
}

@article{huang2025diffusecloc,
  title     = {{Diffuse-CLoC}: Guided Diffusion for Physics-based Character Look-ahead Control},
  author    = {Huang, Xiaoyu and Truong, Takara and Zhang, Yunbo and Yu, Fangzhou and Sleiman, Jean Pierre and Hodgins, Jessica and Sreenath, Koushil and Farshidian, Farbod},
  journal   = {ACM Transactions on Graphics},
  volume    = {44},
  number    = {4},
  pages     = {132:1--132:12},
  articleno = {132},
  publisher = {Association for Computing Machinery (ACM)},
  year      = {2025},
  month     = {July},
  doi       = {10.1145/3731206},
  url       = {https://doi.org/10.1145/3731206}
}

@article{luo2025sonic,
  title     = {{SONIC}: Supersizing motion tracking for natural humanoid whole-body control},
  author    = {Luo, Zhengyi and Yuan, Ye and Wang, Tingwu and Li, Chenran and Casta{\~n}eda, Fernando and Chen, Sirui and Cao, Zi-Ang and Li, Jiefeng and Minor, David and Ben, Qingwei and Park, Jinhyung and Sami, David and Wang, Zi and Da, Xingye and Ding, Runyu and Hogg, Cyrus and Song, Lina and Lim, Edy and Jeong, Eugene and He, Tairan and Xue, Haoru and Xiao, Wenli and Yuen, Simon and Kautz, Jan and Chang, Yan and Iqbal, Umar and Fan, Linxi ``Jim'' and Zhu, Yuke},
  journal   = {Science Robotics},
  volume    = {11},
  number    = {117},
  publisher = {American Association for the Advancement of Science (AAAS)},
  year      = {2026},
  month     = {Aug},
  doi       = {10.1126/scirobotics.aed4592},
  url       = {https://doi.org/10.1126/scirobotics.aed4592}
}

@inproceedings{li2025bfmzero,
  title     = {{BFM-Zero}: A Promptable Behavioral Foundation Model for Humanoid Control Using Unsupervised Reinforcement Learning},
  author    = {Li, Yitang and Luo, Zhengyi and Zhang, Tonghe and Dai, Cunxi and Kanervisto, Anssi and Tirinzoni, Andrea and Weng, Haoyang and Kitani, Kris and Guzek, Mateusz and Touati, Ahmed and Lazaric, Alessandro and Pirotta, Matteo and Shi, Guanya},
  booktitle = {International Conference on Learning Representations},
  editor    = {C. Vondrick and B. Hariharan and C. Raffel and L. Pinto and D. Yang and A. Faust},
  pages     = {79697--79725},
  volume    = {2026},
  year      = {2026},
  url       = {https://proceedings.iclr.cc/paper_files/paper/2026/file/80b677bb4d17f77220be88e1d1fdebde-Paper-Conference.pdf}
}

@misc{zeng2025bfm,
  title         = {Behavior Foundation Model for Humanoid Robots},
  author        = {Zeng, Weishuai and Lu, Shunlin and Yin, Kangning and Niu, Xiaojie and Dai, Minyue and Wang, Jingbo and Pang, Jiangmiao},
  year          = {2025},
  doi           = {10.48550/arxiv.2509.13780},
  eprint        = {2509.13780},
  archiveprefix = {arXiv},
  url           = {https://arxiv.org/abs/2509.13780}
}

@inproceedings{shao2025langwbc,
  title     = {{LangWBC}: Language-directed Humanoid Whole-Body Control via End-to-end Learning},
  author    = {Shao, Yiyang and Zhang, Bike and Liao, Qiayuan and Huang, Xiaoyu and Gao, Yuman and Chi, Yufeng and Li, Zhongyu and Shao, Sophia and Sreenath, Koushil},
  booktitle = {Robotics: Science and Systems XXI},
  series    = {RSS2025},
  publisher = {Robotics: Science and Systems Foundation},
  year      = {2025},
  month     = {June},
  doi       = {10.15607/rss.2025.xxi.065},
  url       = {https://doi.org/10.15607/rss.2025.xxi.065}
}

@misc{robotparty2026mimiclite,
  title        = {{MimicLite}: Efficient and Effective General Humanoid Motion Tracking},
  author       = {{RoboParty Lab Team}},
  year         = {2026},
  howpublished = {\url{https://github.com/EGalahad/mimic-lite}},
  note         = {Technical report: \url{https://github.com/Roboparty/MimicLite/blob/main/mimic-lite.pdf}}
}

@misc{chen2026holomotion,
  title         = {{HoloMotion-1} Technical Report},
  author        = {Chen, Maiyue and Wang, Kaihui and Zhang, Bo and Ma, Xihan and Yang, Zhiyuan and Ren, Yi and Huang, Qijun and Zhu, Zihao and Wang, Yucheng and Su, Zhizhong},
  year          = {2026},
  doi           = {10.48550/arxiv.2605.15336},
  eprint        = {2605.15336},
  archiveprefix = {arXiv},
  url           = {https://arxiv.org/abs/2605.15336}
}

@inproceedings{zhang2025any2track,
  title     = {Track Any Motions under Any Disturbances},
  author    = {Zhang, Zhikai and Guo, Jun and Chen, Chao and Wang, Jilong and Lin, Chenghuai and Lian, Yunrui and Xue, Han and Wang, Zhenrong and Liu, Maoqi and Lyu, Jiangran and Liu, Huaping and Wang, He and Yi, Li},
  booktitle = {2026 IEEE International Conference on Robotics and Automation (ICRA)},
  publisher = {IEEE},
  year      = {2026},
  url       = {https://ras.papercept.net/conferences/conferences/ICRA26/program/ICRA26_ContentListWeb_5.html}
}

@misc{tessler2025protomotions,
  title        = {{ProtoMotions3}: An Open-source Framework for Humanoid Simulation and Control},
  author       = {Tessler, Chen and Jiang, Yifeng and Peng, Xue Bin and Coumans, Erwin and Shi, Yi and Zhang, Haotian and Rempe, Davis and Chechik, Gal and Fidler, Sanja},
  journal      = {GitHub repository},
  publisher    = {GitHub},
  year         = {2025},
  url          = {https://github.com/NVLabs/ProtoMotions},
  howpublished = {\url{https://github.com/NVLabs/ProtoMotions/}}
}

@inproceedings{li2025clone,
  title     = {{CLONE}: Closed-Loop Whole-Body Humanoid Teleoperation for Long-Horizon Tasks},
  author    = {Li, Yixuan and Lin, Yutang and Cui, Jieming and Liu, Tengyu and Liang, Wei and Zhu, Yixin and Huang, Siyuan},
  booktitle = {Proceedings of The 9th Conference on Robot Learning},
  editor    = {Lim, Joseph and Song, Shuran and Park, Hae-Won},
  series    = {Proceedings of Machine Learning Research},
  volume    = {305},
  pages     = {4493--4505},
  month     = {27--30 Sep},
  publisher = {PMLR},
  year      = {2025},
  url       = {https://proceedings.mlr.press/v305/li25h.html}
}

@inproceedings{sferrazza2024humanoidbench,
  title     = {{HumanoidBench}: Simulated Humanoid Benchmark for Whole-Body Locomotion and Manipulation},
  author    = {Sferrazza, Carmelo and Huang, Dun-Ming and Lin, Xingyu and Lee, Youngwoon and Abbeel, Pieter},
  booktitle = {Robotics: Science and Systems XX},
  series    = {RSS2024},
  publisher = {Robotics: Science and Systems Foundation},
  year      = {2024},
  month     = {July},
  doi       = {10.15607/rss.2024.xx.061},
  url       = {https://doi.org/10.15607/rss.2024.xx.061}
}

@article{zhao2026ardy,
  title     = {{ARDY}: Autoregressive Diffusion with Hybrid Representation for Interactive Human Motion Generation},
  author    = {Zhao, Kaifeng and Petrovich, Mathis and Zhang, Haotian and Wang, Tingwu and Tang, Siyu and Rempe, Davis},
  journal   = {ACM Transactions on Graphics (TOG)},
  year      = {2026},
  volume    = {45},
  number    = {4},
  articleno = {86},
  doi       = {10.1145/3811284},
  url       = {https://doi.org/10.1145/3811284}
}

@article{wang2026motionbricks,
  title     = {{MotionBricks}: Scalable Real-Time Motions with Modular Latent Generative Model and Smart Primitives},
  author    = {Wang, Tingwu and Dionne, Olivier and De Ruyter, Michael and Minor, David and Rempe, Davis and Zhao, Kaifeng and Petrovich, Mathis and Yuan, Ye and Li, Chenran and Luo, Zhengyi and Robison, Brian and Blackwell, Xavier and Antoniazzi, Bernardo and Peng, Xue Bin and Zhu, Yuke and Yuen, Simon},
  journal   = {ACM Transactions on Graphics},
  volume    = {45},
  number    = {4},
  pages     = {1--22},
  publisher = {Association for Computing Machinery (ACM)},
  year      = {2026},
  month     = {July},
  doi       = {10.1145/3811334},
  url       = {https://doi.org/10.1145/3811334}
}

@misc{carroll2026scdp,
  title         = {{SCDP}: Learning Humanoid Locomotion from Partial Observations via Mixed-Observation Distillation},
  author        = {Carroll, Milo and Peng, Tianhu and Bao, Lingfan and Zhou, Chengxu and Li, Zhibin},
  year          = {2026},
  doi           = {10.48550/arxiv.2603.09574},
  eprint        = {2603.09574},
  archiveprefix = {arXiv},
  url           = {https://arxiv.org/abs/2603.09574}
}

@misc{jiang2024textop,
  title         = {{TextOp}: Real-time Interactive Text-Driven Humanoid Robot Motion Generation and Control},
  author        = {Xie, Weiji and Zheng, Jiakun and Han, Jinrui and Shi, Jiyuan and Zhang, Weinan and Bai, Chenjia and Li, Xuelong},
  year          = {2026},
  doi           = {10.48550/arxiv.2602.07439},
  eprint        = {2602.07439},
  archiveprefix = {arXiv},
  url           = {https://arxiv.org/abs/2602.07439}
}

@inproceedings{jiang2024hgap,
  title     = {{H-GAP}: Humanoid Control with a Generalist Planner},
  author    = {Jiang, Zhengyao and Xu, Yingchen and Wagener, Nolan and Luo, Yicheng and Janner, Michael and Grefenstette, Edward and Rockt{\"a}schel, Tim and Tian, Yuandong},
  booktitle = {International Conference on Learning Representations},
  volume    = {2024},
  pages     = {17774--17791},
  year      = {2024},
  url       = {https://openreview.net/forum?id=LYG6tBlEX0}
}

@article{tessler2024maskedmimic,
  title     = {{MaskedMimic}: Unified Physics-Based Character Control Through Masked Motion Inpainting},
  author    = {Tessler, Chen and Guo, Yunrong and Nabati, Ofir and Chechik, Gal and Peng, Xue Bin},
  journal   = {ACM Transactions on Graphics},
  volume    = {43},
  number    = {6},
  pages     = {1--21},
  publisher = {Association for Computing Machinery (ACM)},
  year      = {2024},
  month     = {Nov},
  doi       = {10.1145/3687951},
  url       = {https://doi.org/10.1145/3687951}
}

@misc{zhang2026script,
  title         = {{SCRIPT}: Scalable Diffusion Policy with Multi-stage Training for Language-driven Physics-Based Humanoid Control},
  author        = {Zhang, Jingyan and Liang, Han and Zhang, Ruichi and Li, Bin and Zhang, Juze and Chen, Xin and Wang, Jingya and Xu, Lan and Yu, Jingyi},
  year          = {2026},
  doi           = {10.48550/arxiv.2605.22894},
  eprint        = {2605.22894},
  archiveprefix = {arXiv},
  url           = {https://arxiv.org/abs/2605.22894}
}

@article{gu2026refinedp,
  title     = {{REFINE-DP}: Diffusion Policy Fine-Tuning for Humanoid Loco-Manipulation via Reinforcement Learning},
  author    = {Gu, Zhaoyuan and Chen, Yipu and Chai, Zimeng and Cueva, Alfred and Nguyen, Thong and Wu, Yifan and Xue, Huishu and Kim, Minji and Legene, Isaac and Liu, Fukang and Kim, KyoungMok and Barula, Ayan and Chen, Yongxin and Zhao, Ye},
  journal   = {IEEE Robotics and Automation Letters},
  volume    = {11},
  number    = {10},
  pages     = {11617--11624},
  publisher = {Institute of Electrical and Electronics Engineers (IEEE)},
  year      = {2026},
  month     = {Oct},
  doi       = {10.1109/lra.2026.3723742},
  url       = {https://doi.org/10.1109/lra.2026.3723742}
}

@inproceedings{lu2022dpm,
  title     = {DPM-Solver: A Fast {ODE} Solver for Diffusion Probabilistic Model Sampling in Around 10 Steps},
  author    = {Lu, Cheng and Zhou, Yuhao and Bao, Fan and Chen, Jianfei and Li, Chongxuan and Zhu, Jun},
  booktitle = {Advances in Neural Information Processing Systems 35},
  series    = {NeurIPS 2022},
  pages     = {5775--5787},
  publisher = {Neural Information Processing Systems Foundation, Inc. (NeurIPS)},
  year      = {2022},
  doi       = {10.52202/068431-0418},
  url       = {https://doi.org/10.52202/068431-0418}
}

@inproceedings{ross2011reduction,
  title     = {A Reduction of Imitation Learning and Structured Prediction to No-Regret Online Learning},
  author    = {Ross, Stephane and Gordon, Geoffrey and Bagnell, Drew},
  booktitle = {Proceedings of the Fourteenth International Conference on Artificial Intelligence and Statistics},
  series    = {Proceedings of Machine Learning Research},
  volume    = {15},
  pages     = {627--635},
  publisher = {PMLR},
  year      = {2011},
  month     = {11--13 Apr},
  url       = {https://proceedings.mlr.press/v15/ross11a.html},
  address   = {Fort Lauderdale, FL, USA}
}

@inproceedings{laskey2017dart,
  title     = {{DART}: Noise Injection for Robust Imitation Learning},
  author    = {Laskey, Michael and Lee, Jonathan and Fox, Roy and Dragan, Anca and Goldberg, Ken},
  booktitle = {Proceedings of the 1st Annual Conference on Robot Learning},
  series    = {Proceedings of Machine Learning Research},
  volume    = {78},
  pages     = {143--156},
  publisher = {PMLR},
  year      = {2017},
  month     = {13--15 Nov},
  url       = {https://proceedings.mlr.press/v78/laskey17a.html}
}

@inproceedings{kelly2019hg,
  title     = {{HG-DAgger}: Interactive Imitation Learning with Human Experts},
  author    = {Kelly, Michael and Sidrane, Chelsea and Driggs-Campbell, Katherine and Kochenderfer, Mykel J.},
  booktitle = {2019 International Conference on Robotics and Automation (ICRA)},
  pages     = {8077--8083},
  publisher = {IEEE},
  year      = {2019},
  month     = {May},
  doi       = {10.1109/icra.2019.8793698},
  url       = {https://doi.org/10.1109/icra.2019.8793698}
}

@inproceedings{2022Song_DDIM,
  title         = {Denoising Diffusion Implicit Models},
  author        = {Jiaming Song and Chenlin Meng and Stefano Ermon},
  booktitle     = {International Conference on Learning Representations},
  year          = {2021},
  url           = {https://openreview.net/forum?id=St1giarCHLP}
}

@inproceedings{2023Song_ConsistencyModels,
  title     = {Consistency Models},
  author    = {Song, Yang and Dhariwal, Prafulla and Chen, Mark and Sutskever, Ilya},
  booktitle = {Proceedings of the 40th International Conference on Machine Learning},
  series    = {Proceedings of Machine Learning Research},
  volume    = {202},
  pages     = {32211--32252},
  publisher = {PMLR},
  year      = {2023},
  month     = {23--29 Jul},
  url       = {https://proceedings.mlr.press/v202/song23a.html}
}

@inproceedings{2024Ren_DPPO,
  title     = {Diffusion Policy Policy Optimization},
  author    = {Ren, Allen Z. and Lidard, Justin and Ankile, Lars L. and Simeonov, Anthony and Agrawal, Pulkit and Majumdar, Anirudha and Burchfiel, Benjamin and Dai, Hongkai and Simchowitz, Max},
  booktitle = {International Conference on Learning Representations},
  volume    = {2025},
  pages     = {77288--77329},
  year      = {2025},
  url       = {https://openreview.net/forum?id=mEpqHvbD2h}
}

@misc{chen2026reactivebfm,
  title         = {{ReactiveBFM}: Reactive Closed-Loop Motion Planning Towards Universal Humanoid Whole-Body Control},
  author        = {Chen, Xiao and Zeng, Weishuai and Niu, Xiaojie and Wang, Zirui and Li, Jianan and Wang, Huayi and Xu, Furui and Chen, Jiahe and Zhong, Weixiang and Ding, Lihe and Li, Kailin and Pang, Jiangmiao and Wang, Tai and Xue, Tianfan and Wang, Jingbo},
  year          = {2026},
  doi           = {10.48550/arxiv.2606.30362},
  eprint        = {2606.30362},
  archiveprefix = {arXiv},
  url           = {https://arxiv.org/abs/2606.30362}
}

@inproceedings{wu2025uniphys,
  title     = {{UniPhys}: Unified Planner and Controller with Diffusion for Flexible Physics-Based Character Control},
  author    = {Wu, Yan and Karunratanakul, Korrawe and Luo, Zhengyi and Tang, Siyu},
  booktitle = {Proceedings of the IEEE/CVF International Conference on Computer Vision (ICCV)},
  pages     = {13214--13224},
  year      = {2025},
  month     = {Oct}
}

@inproceedings{li2026roboghost,
  title     = {From Language to Locomotion: Retargeting-free Humanoid Control via Motion Latent Guidance},
  author    = {Li, Zhe and Wei, Yangyang and Zhu, Boan and Peng, Yibo and Huang, Tao and Wang, Pengwei and Wang, Zhongyuan and Chi, Cheng and Xu, Chang and Zhang, Shanghang},
  booktitle = {International Conference on Learning Representations},
  year      = {2026}
}

@inproceedings{wang2026sentinel,
  title     = {End-to-End Language-Action Model for Humanoid Whole Body Control},
  author    = {Wang, Yuxuan and Jiang, Haobin and Yao, Shiqing and Ding, Ziluo and Lu, Zongqing},
  booktitle = {Proceedings of the IEEE/CVF Conference on Computer Vision and Pattern Recognition (CVPR)},
  pages     = {38399--38409},
  year      = {2026},
  month     = {June}
}

@inproceedings{chen2025falcon,
  title     = {{Falcon}: Fast Visuomotor Policies via Partial Denoising},
  author    = {Chen, Haojun and Liu, Minghao and Ma, Chengdong and Ma, Xiaojian and Ma, Zailin and Wu, Huimin and Chen, Yuanpei and Zhong, Yifan and Wang, Mingzhi and Li, Qing and Yang, Yaodong},
  booktitle = {Proceedings of the 42nd International Conference on Machine Learning},
  series    = {Proceedings of Machine Learning Research},
  volume    = {267},
  pages     = {8552--8573},
  publisher = {PMLR},
  year      = {2025},
  month     = {13--19 Jul}
}

@inproceedings{prasad2024consistencypolicy,
  title     = {Consistency Policy: Accelerated Visuomotor Policies via Consistency Distillation},
  author    = {Prasad, Aaditya and Lin, Kevin and Wu, Jimmy and Zhou, Linqi and Bohg, Jeannette},
  booktitle = {Proceedings of Robotics: Science and Systems},
  year      = {2024},
  month     = {Jul},
  address   = {Delft, Netherlands},
  doi       = {10.15607/RSS.2024.XX.071}
}
\clearpage
\appendix
\section{Extended Related-Work Coverage}
\label{app:extended_related_work}

\subsection{Deployment Hardware and Timing Sources}
\label{app:deployment_comparison}

\begin{table*}[!htbp]
  \centering
  \caption{Policy capabilities and deployment configurations of humanoid and legged generative control systems. Top: representation and task interfaces. Bottom: inference hardware, location, and reported execution rates; component timings are not complete-loop measurements.}
  \label{tab:capability-deployment-full}
  \setlength{\tabcolsep}{2.5pt}
  \renewcommand{\arraystretch}{1.10}
  \scriptsize
  \begin{tabular*}{\textwidth}{@{\extracolsep{\fill}}llcccccccc@{}}
    \toprule
    & \multicolumn{4}{c}{Model settings}
    & \multicolumn{4}{c}{Task capability}
    & \multicolumn{1}{c}{Deployment} \\
    \cmidrule(lr){2-5}\cmidrule(lr){6-9}\cmidrule(lr){10-10}
    Method
    & Paradigm
    & Future pred.
    & Proprio.
    & Rolling reuse
    & CFG
    & CG
    & Text
    & Joy/Goal
    & Robot \\
    \midrule
    MotionBricks + SONIC & Generator + tracker & $\diamond$ & $\diamond$ & -- & -- & -- & -- & \cmark & \cmark \\
    ARDY + SONIC & Open-loop track & $\diamond$ & $\diamond$ & -- & $\diamond$ & -- & $\diamond$ & \cmark & -- \\
    TextOp & Open-loop track & $\diamond$ & $\diamond$ & -- & \cmark & -- & \cmark & \cmark & \cmark \\
    ReactiveBFM & Closed-loop track & $\diamond$ & $\diamond$ & -- & -- & -- & \cmark & \cmark & \cmark \\
    CLoSD & Closed-loop track & $\diamond$ & -- & -- & \cmark & -- & \cmark & \cmark & -- \\
    DiffuseLoco & Action diffusion & \cmark & \cmark & -- & -- & -- & -- & \cmark & \cmark \\
    RoboGhost & Latent gen. + diff. tracker & $\diamond$ & \cmark & -- & -- & -- & \cmark & -- & \cmark \\
    SENTINEL & Action flow + residual RL & \cmark & \cmark & -- & \cmark & -- & \cmark & \cmark & \cmark \\
    Diffuse-CLoC & Joint diffusion & \cmark & -- & \cmark & -- & \cmark & -- & \cmark & -- \\
    BeyondMimic & Latent joint diffusion + dec. & \cmark & $\diamond$ & -- & -- & \cmark & -- & \cmark & \cmark \\
    SCDP & Joint diffusion & \cmark & \cmark & -- & -- & -- & -- & \cmark & \cmark \\
    SCRIPT & Joint diffusion & \cmark & -- & -- & \cmark & -- & \cmark & \cmark & -- \\
    UniPhys & Latent joint diffusion + dec. & \cmark & -- & -- & \cmark & \cmark & \cmark & \cmark & -- \\
    \midrule
    \textbf{PredActor} & Joint diffusion & \cmark & \cmark & \cmark & \cmark & \cmark & \cmark & \cmark & \cmark \\
    \bottomrule
  \end{tabular*}
  \par\medskip
  \begin{tabular}{@{}>{\raggedright\arraybackslash}p{0.18\textwidth}>{\raggedright\arraybackslash}p{0.14\textwidth}>{\raggedright\arraybackslash}p{0.15\textwidth}>{\raggedright\arraybackslash}p{0.17\textwidth}>{\raggedright\arraybackslash}p{0.29\textwidth}@{}}
    \toprule
    Method & Inference CPU & Inference GPU & Compute location & Rate / timing (reported boundary) \\
    \midrule
    MotionBricks + SONIC & Orin SoC$^{a}$ & Orin integrated$^{a}$ & Fully onboard G1 & 50 Hz tracker$^{*}$; 10 Hz replanning; 5 ms generator \\
    ARDY + SONIC & -- & RTX 4090$^{b}$ & Workstation$^{b}$ & 33 / 63 ms generator (4 / 10 steps)$^{b}$ \\
    TextOp & i7-13700$^{c}$ & RTX 4090$^{c}$ & External generator; onboard tracker & 6.25 Hz generator; 50 Hz tracker \\
    ReactiveBFM & i9-13900K & RTX 4090 & Offboard workstation & 50 Hz controller; async planner (19.3 ms) \\
    CLoSD & -- & -- & Simulation only & 11.4 ms for a 2 s motion plan \\
    DiffuseLoco & i7-13700H & RTX 4060 Mobile & Onboard Go1 mini-PC & 30 Hz control target; 116.5 Hz inference \\
    RoboGhost & Orin NX SoC$^{d}$ & Orin NX integrated$^{d}$ & Partially onboard G1$^{d}$ & 50 Hz one-step policy; DDIM-2 4.8 ms; 18--30 ms transmission \\
    SENTINEL & -- & RTX 4090 & Offboard (wired) & 50 Hz policy; 200 Hz PD; 16 ms (5 passes) + 2 ms communication \\
    Diffuse-CLoC & -- & RTX 4060 & Simulation only & 77.41 Hz inference (selected profile) \\
    BeyondMimic & -- (CPU decoder) & RTX 4060 Mobile & Onboard mini-PC (v4) & 25 Hz control; async diffusion about 20 ms \\
    SCDP & -- & RTX 5090 & Remote workstation & 50 Hz control; 105 Hz inference \\
    SCRIPT & -- & -- & Simulation only & Physical control: N/A \\
    UniPhys & -- & NVIDIA A100 & Simulation only & 9.5--23 FPS (5-step DDIM; schedule dependent) \\
    \midrule
    \textbf{PredActor} & \textbf{Orin NX SoC} & \textbf{Orin NX integrated} & \textbf{Fully onboard G1} & \textbf{50 Hz control; callback p95 19.383 ms} \\
    \bottomrule
  \end{tabular}
  \vspace{2pt}

  \parbox{0.985\textwidth}{\scriptsize
  \textbf{Symbols:} \cmark: supported; $\diamond$: partial/modular; $\times$: explicitly unsupported; --: unreported/unverified; N/A: not applicable.\par
  \textbf{Capability definitions:} Paradigm describes the deployed data path rather than an author's end-to-end label. RoboGhost uses a separate future motion-latent generator followed by a one-step, 23-DoF target-joint diffusion tracker. SENTINEL's flow model predicts future 29-DoF joint-position actions from language and proprioceptive history. These actions are passed to PD rather than serving as an intermediate kinematic reference; a separately PPO-trained residual policy corrects them during physical deployment. Future pred. means that the robot-facing generative policy outputs multiple future control/state steps; $\diamond$ means that only a separate planner/generator does so. Proprio. means proprioceptive input; dec. means decoder; Robot denotes physical deployment. Proprioceptive input permits task commands but excludes privileged physical-state input; $\diamond$ marks input inherited from another module, such as a tracker in a generator--tracker system. Rolling reuse requires explicit reuse of overlapping state/action predictions or their partially denoised representations, not just replanning. CFG requires conditional--null prediction mixing; $\diamond$ marks ARDY generator-level CFG/text support without separate validation in the SONIC composition.\par
  \textbf{Hardware notes:} $^{*}$50 Hz is sourced from MotionBricks' adopted SONIC tracker, not independently reported for the composition. $^{a}$MotionBricks places its exported models on Jetson Orin and discusses resource competition with the low-level controller, supporting fully onboard composition; the Orin variant, CPU model, and full-stack timing are unreported. $^{b}$ARDY generator-demo benchmark; robot-stack hardware and rate are unreported. $^{c}$TextOp external workstation; tracker CPU is unreported. $^{d}$RoboGhost establishes only the 50 Hz control policy on an onboard Orin NX; the compute location of its upstream text encoder and motion generator is unspecified, and 4.8 ms covers the student policy. Inference hardware is not inferred from training hardware.\par
  \textbf{Timing boundary:} PredActor's callback covers received robot-state messages through host-action production; 593/600 calls meet 20 ms. The timing benchmark does not actuate the robot. Component timings are not full-loop measurements. Row-level source evidence follows below.}
\end{table*}

Table~\ref{tab:capability-deployment-full} expands the main-text comparison with the full capability matrix, inference hardware, and timing boundaries. It separates inference hardware from training hardware and robot control rates from generator throughput. MotionBricks reports deploying its exported models on the G1's Jetson Orin with 10 Hz replanning and 5 ms generator latency (\href{https://arxiv.org/html/2604.24833v1}{Section 7.4})~\citep{wang2026motionbricks}. Its discussion of avoiding resource competition with the low-level controller supports a fully onboard generator--tracker composition; the Orin variant, processor assignments, and complete-loop latency remain unspecified. The 50 Hz tracker rate comes from the adopted SONIC controller (\href{https://arxiv.org/html/2511.07820v1}{Section 3.5})~\citep{luo2025sonic}. ARDY reports an RTX 4090 interactive-generator benchmark of 33/63 ms for 4/10 diffusion steps (\href{https://arxiv.org/html/2607.08741v1}{Section 4.2})~\citep{zhao2026ardy}; robot-stack hardware and rate are not reported.

The two generator--tracker compositions have separate capability rows. MotionBricks steers motion through keyframes and root-trajectory constraints supplied by smart primitives (Sections 5--6). Its Section 7.2 CFG discussion concerns comparison baselines; the masked-token generator is reported without conditional--null prediction mixing or sampling-time objective gradients. ARDY explicitly supports text and spatial conditioning with CFG (Section 3.5), recorded as generator-level capabilities in the SONIC composition. The ballet and locomotion videos under \href{https://research.nvidia.com/labs/sil/projects/ardy/}{Humanoid Robot Control on its official project page} show simulated G1 tracking, so the table records simulation rather than physical deployment. Spatial steering and CG remain separate columns, and the proprioceptive-input marks refer to the trackers.

TextOp runs its generator on an external i7-13700/RTX 4090 workstation at 6.25 Hz and its tracker onboard at 50 Hz (\href{https://arxiv.org/html/2602.07439v1}{Section III-E and Appendix B-F})~\citep{jiang2024textop}. ReactiveBFM uses an offboard i9-13900K/RTX 4090 workstation, a 50 Hz controller, and an asynchronous planner; the reported 19.3 ms planner and 5.9 ms controller timings are component measurements (\href{https://arxiv.org/html/2606.30362v1}{Deployment and Appendix B.5})~\citep{chen2026reactivebfm}. SCDP uses a remote RTX 5090 workstation for 50 Hz G1 control and reports 105 Hz inference throughput (\href{https://arxiv.org/html/2603.09574v1}{Section II-D})~\citep{carroll2026scdp}.

RoboGhost is hierarchical at the system level: a separate autoregressive motion generator predicts a future motion-latent sequence from text, which conditions a proprioceptive diffusion controller. The lower controller denoises one 23-dimensional target-joint command per control tick rather than a future action horizon. Its future-prediction entry is therefore modular ($\diamond$): the upstream generator predicts future motion, whereas the robot-facing controller outputs one control step. It is trained by DAgger-style imitation of an RL teacher that tracks explicit reference motion, although deployment bypasses explicit motion decoding and retargeting. The deployment section places the 50~Hz control policy on a Jetson Orin NX and reports a 500~Hz low-level controller, 4.8~ms two-step DDIM policy time, and 18--30~ms command transmission (\href{https://proceedings.iclr.cc/paper_files/paper/2026/file/d1ef0be5d288aba023664aa96a510486-Paper-Conference.pdf}{Sections 3.3 and 9.4; Table 14})~\citep{li2026roboghost}. The upstream text encoder and motion generator have unspecified compute placement, and the \href{https://github.com/gentlefress/RoboGhost}{released real-robot script} loads a precomputed motion-latent file. The table therefore records the full text-to-action stack as partially onboard.

SENTINEL's flow model predicts an $H$-step chunk whose entries are linearly projected to 29-DoF target joint positions. These are joint-position actions passed to joint-space PD, not an intermediate root/keypoint reference trajectory or direct motor torques. The action model is closed-loop: its language--state encoder conditions generation on recent 50~Hz proprioceptive history and a 4~Hz history spanning 10~s, and replans in a receding-horizon manner. This places the base model in action-flow generative control. Physical deployment additionally applies a residual head trained separately with PPO, yielding $a_t+\Delta a_t$ before the fixed 200~Hz PD controller. The base-versus-residual ablation is conducted in simulation, while the physical G1 evaluation uses the model after residual post-training; the deployed stack is therefore modular, while the generated variable remains an action. The flow model runs on an external wired RTX~4090; five passes take 16~ms, communication adds about 2~ms, and states arrive at 50~Hz (\href{https://arxiv.org/pdf/2511.19236}{Sections 4.2, 5.4--5.6, and C.4; Table 11})~\citep{wang2026sentinel}. Its asynchronous chunk replacement is classified as replanning rather than rolling reuse.

For BeyondMimic, Table~\ref{tab:capability-deployment-full} uses \href{https://arxiv.org/html/2508.08241v4}{arXiv v4, Supplement S4 (Diffusion)}: the diffusion-stage setup runs at 25 Hz, with onboard RTX 4060 Mobile inference, a synchronous CPU VAE decoder, and approximately 20 ms asynchronous diffusion latency~\citep{liao2025beyondmimic}. Earlier versions (\href{https://arxiv.org/html/2508.08241v1}{v1}, \href{https://arxiv.org/html/2508.08241v2}{v2}, and \href{https://arxiv.org/html/2508.08241v3}{v3}, Section VI-B) describe offboard inference and omit this setting. Its latent state--action model is classified as joint diffusion rather than a kinematic generator followed by a reference tracker.

DiffuseLoco reports onboard Go1 deployment on an i7-13700H/RTX 4060 Mobile mini-PC; its 30 Hz control requirement differs from the 116.5 Hz TensorRT inference benchmark (\href{https://arxiv.org/html/2404.19264v1}{Appendix E and Figure 16})~\citep{huang2024diffuseloco}. Diffuse-CLoC evaluates simulated characters and reports RTX 4060 inference, with 77.41 Hz for the selected profile (\href{https://arxiv.org/html/2503.11801v2}{Section V-B and Table II})~\citep{huang2025diffusecloc}. CLoSD's 11.4 ms measurement generates a two-second motion plan, not a physical-robot callback (\href{https://xbpeng.github.io/projects/CLoSD/CLoSD_2025.pdf}{Section 4.1})~\citep{tevet2024closd}; SCRIPT evaluates simulated humanoids rather than onboard robot execution (\href{https://arxiv.org/html/2605.22894v1}{Section 5.1})~\citep{zhang2026script}. Unreported hardware and rates are marked ``--'', even when training hardware is stated; N/A denotes physical control rates for simulation-only systems.

UniPhys jointly diffuses canonicalized simulator states and latent actions decoded by PULSE, and supports both conditional--null text CFG and objective-gradient guidance. Because its state contains root pose and velocity and its evaluation is simulation-only, it is not marked as proprioceptive robot deployment. With five-step DDIM on an NVIDIA A100, the reported schedules run at 9.5--23 FPS (\href{https://openaccess.thecvf.com/content/ICCV2025/html/Wu_UniPhys_Unified_Planner_and_Controller_with_Diffusion_for_Flexible_Physics-Based_ICCV_2025_paper.html}{Sections 4.2--5.5 and Table 4})~\citep{wu2025uniphys}. Falcon is a training-free add-on that initializes a new action horizon from selected partially denoised actions stored from earlier control ticks; its real dual-arm tasks report 0.14 and 0.15 s for Falcon+DDPM (\href{https://proceedings.mlr.press/v267/chen25ao.html}{Sections 4.5 and A.2; Tables 5--6})~\citep{chen2025falcon}. Consistency Policy instead distills Diffusion Policy into a one- or few-evaluation action generator, with real-robot cleanup and insertion policies running on an external RTX 3070 Ti laptop at 21--22 ms (\href{https://www.roboticsproceedings.org/rss20/p071.html}{Section IV-C and Table IV})~\citep{prasad2024consistencypolicy}. Both are manipulation-focused acceleration references and are therefore discussed here but excluded from the humanoid/legged comparison tables.

PredActor combines proprioceptive conditioning, CG/CFG, direct joint-action execution, and fully onboard G1 inference on Jetson Orin NX. Its measured callback runs from received robot-state messages to host-action production, with p50/p95 of 16.790/19.383 ms and 593/600 calls within 20 ms. Physical execution is demonstrated separately; the timing benchmark issues no actuation commands.

\subsection{Motion-Control Landscape}
\label{app:motion_control_landscape}

The literature spans several control paradigms that are easy to conflate. A
\emph{motion tracker} receives a reference trajectory or motion representation
and produces physically executed actions. PHC established scalable perpetual tracking and recovery for simulated avatars~\cite{luo2023perpetual}. Recent robot trackers emphasize different bottlenecks: MimicLite reduces the reported training cost of a deployable G1 tracker~\cite{robotparty2026mimiclite}; SONIC, released as GEAR-SONIC, scales model capacity, motion data, and compute while exposing several downstream motion interfaces~\cite{luo2025sonic}; HoloMotion-1 uses a hybrid motion corpus and temporal mixture-of-experts model for zero-shot whole-body tracking~\cite{chen2026holomotion}; and Any2Track couples a general tracker with a history-conditioned dynamics adapter for disturbances~\cite{zhang2025any2track}. ProtoMotions3 is instead an open simulation, learning, and deployment framework that can train tracking policies; it is not one additional controller architecture~\cite{tessler2025protomotions}. These trackers and tools strengthen the execution side of generator--tracker stacks, but do not themselves select a long-horizon task behavior without a reference or higher-level input.

\subsection{Reactive Planner--Tracker Comparisons}
\label{app:reactivebfm_comparison}

ReactiveBFM distinguishes open-loop cascades from closed-loop compositions with physical-state feedback; those labels describe the evaluated composition and do not imply that every component replans from physical state ~\citep{chen2026reactivebfm}.

A second family broadens the policy interface. BFM-Zero learns a shared latent space for motion, goal, and reward prompts using unsupervised reinforcement learning~\cite{li2025bfmzero}. BFM4Humanoid combines masked online distillation with a conditional variational autoencoder to support multiple whole-body control modes~\cite{zeng2025bfm}. LangWBC is a non-diffusion comparison: reinforcement learning, policy distillation, and a conditional variational autoencoder yield one language-conditioned policy that outputs actions directly~\cite{shao2025langwbc}. These are unified, promptable, or direct controllers rather than a kinematic planner followed by an independently trained tracker.

Teleoperation systems occupy another role. H2O, OmniH2O, TWIST, and CLONE map live human inputs to whole-body robot control, with CLONE specifically adding closed-loop global-position correction for long-horizon teleoperation ~\cite{he2024h2o,he2024omni,ze2024twist,li2025clone}. They are important sources of control interfaces and demonstrations, but are not autonomous motion generators in the sense used here. HumanoidBench, introduced by~\cite{sferrazza2024humanoidbench}, is likewise an evaluation suite for locomotion and whole-body manipulation rather than a controller. This taxonomy motivates comparisons along reference dependence, task conditioning, generated variables, deploy-time observations, guidance, and closed-loop compute rather than a single overloaded ``end-to-end'' label.

\subsection{Trajectory Generative Models for Control}

Diffusion control methods differ in what they generate. Action-only policies denoise action chunks conditioned on observations~\cite{chi2023diffusion}; this formulation extends to offline legged locomotion~\cite{huang2024diffuseloco} and physics-based character control~\cite{truong2024physics}. Joint trajectory models also represent future states: Diffuser performs return- or constraint-guided planning~\cite{janner2022planning}, while Decision Diffuser uses conditional state-sequence generation with inverse dynamics to recover actions~\cite{ajay2023conditional}. Outside diffusion, H-GAP learns a discrete latent state--action trajectory model and uses model-predictive planning for generalist simulated humanoid control~\cite{jiang2024hgap}. PredActor follows the joint state--action diffusion route: future states support guidance within the model that generates executable actions.

Recent humanoid systems provide closer comparisons. BeyondMimic guides a latent diffusion model of human motion primitives toward test-time objectives and transfers the resulting controller to hardware~\cite{liao2025beyondmimic}. SCDP generates interleaved future states and actions from onboard observation histories, using privileged future trajectories for supervision and partial observations at deployment~\cite{carroll2026scdp}. SCRIPT jointly represents action, state, and text streams in a diffusion transformer and adds reinforcement-learning post-training for language-driven simulated humanoids~\cite{zhang2026script}. Within this landscape, PredActor combines proprioceptive-history conditioning with learned-command CFG and future-state CG, then implements these interfaces within an onboard control loop. Generated variables, policy inputs, guidance mechanisms, and hardware evidence therefore form separate comparison axes in Table~\ref{tab:capability-comparison}.

\subsection{Efficient and Temporally Consistent Diffusion Inference}

Real-time control constrains the cost of iterative sampling. DDIM introduces a non-Markovian implicit process that permits deterministic subsampling without retraining~\cite{2022Song_DDIM}; DPM-Solver uses dedicated high-order integration for diffusion ODEs~\cite{lu2022dpm}; and consistency models learn direct mappings from noise to data~\cite{2023Song_ConsistencyModels}. Robot-specific distillation similarly trades extra training for fewer evaluations: OneDP distills a pretrained diffusion policy into a one-step generator~\cite{wang2025onedp}. These approaches accelerate or replace the sampler, but do not specify how successive control horizons should share computation.

Temporal reuse addresses that second issue. Streaming Diffusion Policy carries forward a partially denoised action trajectory whose near-term entries are cleaner than its tail~\cite{hoeg2024streaming}. TEDi entangles diffusion time with motion time in a recursively advanced buffer for long-horizon motion synthesis~\cite{zhang2024tedi}, while Diffusion Forcing assigns per-token noise levels to combine next-token prediction with full-sequence diffusion~\cite{chen2024diffusion}. Action chunking with temporal ensembling smooths overlapping predictions~\cite{zhao2023learning}, and constrained diffusion adds feasibility projections at sampling time~\cite{romer2025dpcc}. PredActor's $(K,h)$ noise matrix provides one implementation of per-position schedules, DDIM jumps, and rolling-buffer reuse.

\subsection{Imitation Learning and Iterative Data Aggregation}

Behavior cloning suffers from covariate shift because learner actions change the future state distribution. DAgger queries the expert on learner-induced states and aggregates the resulting labels~\cite{ross2011reduction}. DART instead perturbs demonstrations to collect recovery behavior without executing the novice during collection~\cite{laskey2017dart}, whereas HG-DAgger gates human intervention using a learned uncertainty threshold~\cite{kelly2019hg}. These methods separate three design choices that are sometimes conflated: who visits the state, who supplies the label, and whether the collected sample is appended to replay.

Diffusion policies can also be trained with reward optimization rather than label aggregation. DPPO supplies a policy-gradient recipe for fine-tuning diffusion policies~\cite{2024Ren_DPPO}, and REFINE-DP jointly fine-tunes a diffusion planner and low-level controller for humanoid loco-manipulation~\cite{gu2026refinedp}. PredActor instead uses a standard iterative aggregation recipe: learner-rollout origins seed isolated teacher continuations, whose trajectory labels enter combined replay for continued diffusion reconstruction. This training procedure is distinct from offline noise injection and reinforcement-learning fine-tuning.

\subsection{Conditioning and Humanoid Control Interfaces}

Humanoid objectives may be expressed as velocity commands, sparse body targets, motion clips, language, or teleoperation. MotionCLIP aligns text and human motion embeddings~\cite{tevet2022motionclip}; CLoSD couples text-conditioned motion diffusion to physics-based control~\cite{tevet2024closd}; and TextOp streams text changes through a motion generator and a low-level tracking policy~\cite{jiang2024textop}. MaskedMimic treats diverse partial motion descriptions---including text, keyframes, objects, and paths---as masks for one physics-based controller~\cite{tessler2024maskedmimic}. Whole-body teleoperation systems such as H2O, OmniH2O, and TWIST instead map live human motion to humanoid control~\cite{he2024h2o,he2024omni,ze2024twist}.

PredActor exposes each conditioning source through a shared runtime interface. Text embeddings, joystick commands, semantic interpolation, and whole-body guidance can therefore reuse the same exported policy interface. The experiments evaluate behavioral execution and control-period timing separately from interface availability.

\section{Extended Method and System Details}
\label{app:extended_methods}

The following details separate PredActor's predictive representation from its training and execution machinery. Future states are internal variables for guidance, proprioceptive histories supply policy inputs, and selected actions are issued to the joint controller. Teacher supervision, rolling schedules, and runtime optimizations implement this policy without introducing a separate motion-generation and tracking hierarchy at deployment.

\subsection{Predictive State--Action Representation}
\label{app:sec:prdp}

\subsubsection{Proprioceptive inputs and internal future states}

Table~\ref{tab:observation-state-components} separates the deployable observation from the physical state predicted by PredActor. The evaluated PredActor policies use a 96-dimensional observation that includes base linear velocity. The 192-dimensional physical state is mapped by the fixed emphasis projection to a 384-dimensional model representation. The action $\mathbf{a}(t)$ contains 29 joint-controller commands, converted to PD targets by the runtime. Optional task context $\mathbf{z}$ encodes the requested task; the text-conditioned setting uses a 512-dimensional semantic embedding. Normalization statistics are fitted on the training split and fixed at test time.

\begin{table}[h!]
  \centering
  \caption{Observation and physical-state components for the selected G1
  profile. The evaluated 96-dimensional observation includes base linear
  velocity. The emphasis projection changes representation dimension, not the
  underlying physical state.}
  \label{tab:observation-state-components}
  \small
  \setlength{\tabcolsep}{5pt}
  \renewcommand{\arraystretch}{1.04}
  \begin{tabular}{@{}>{\raggedright\arraybackslash}p{0.50\linewidth}r>{\raggedright\arraybackslash}p{0.29\linewidth}@{}}
    \toprule
    Component & Dimension & Description \\
    \midrule
    \multicolumn{3}{@{}l}{\textit{Deployable observation $\mathbf{o}_t$}} \\
    Joint positions & 29 & Robot joint coordinates \\
    Joint velocities & 29 & Robot joint velocities \\
    Projected gravity & 3 & Gravity in the base frame \\
    Base angular velocity & 3 & IMU angular velocity \\
    Previous action & 29 & Last joint command \\
    Base linear velocity & 3 & Base-frame linear velocity \\
    \textbf{Selected observation total} & \textbf{96} & Used by the evaluated profile \\
    \midrule
    \multicolumn{3}{@{}l}{\textit{Physical state $\mathbf{s}(t)$}} \\
    Local body positions & 90 & 30 bodies $\times$ 3 coordinates \\
    Local body linear velocities & 90 & 30 bodies $\times$ 3 velocities \\
    Root position & 3 & Local character frame \\
    Root orientation & 3 & Local rotation representation \\
    Root linear velocity & 3 & Local character frame \\
    Root angular velocity & 3 & Local character frame \\
    \textbf{Physical-state total} & \textbf{192} & $d_s^{\mathrm{phys}}$ \\
    Emphasis-projected state & 384 & Model representation $d_s$ \\
    \bottomrule
  \end{tabular}
\end{table}

The model directly predicts clean trajectories and is optimized with reconstruction losses on both streams. PredActor does not contain a separate deterministic observation-to-state module. In the evaluated setting, deployment initializes the entire state stream from Gaussian noise and reconstructs it jointly with actions through conditional co-diffusion. The reconstructed state trajectory remains internal; only an action is sent to the robot. Onboard signal processing remains part of observation construction.

\subsubsection{Interleaved conditional transformer}

State and action inputs are projected separately, concatenated with sinusoidal embeddings of their respective noise levels, and interleaved as $[s^{\mathrm{tok}}_1,a^{\mathrm{tok}}_1,\ldots,s^{\mathrm{tok}}_h,a^{\mathrm{tok}}_h]$. Learned positional embeddings retain horizon order. A transformer decoder applies self-attention over this sequence and cross-attention to encoded condition tokens. In the reference attention mask, state queries attend to all state tokens but not action tokens; action queries attend causally to state and action tokens. Separate output heads predict clean states and clean actions. The current reference backbone uses two decoder layers, four attention heads, and a 256-dimensional embedding; these are configuration choices rather than architectural requirements.

A fixed, profile-driven state emphasis transform $E$ can be applied before the backbone and its pseudoinverse $E^\dagger$ after prediction. The evaluated transform concatenates a weighted symmetric random projection with an identity component. It is specified before training and is not learned by gradient descent.

\subsection{Dataset Construction and Iterative Data Aggregation}
\label{app:dagger_design}

\subsubsection{Kinematic library and initial rollouts}

Figure~\ref{fig:method_training}(a) combines prompted motion generation and annotated motion sources into a common kinematic library with task labels. A tracking teacher executes these references under observation noise, action noise, and external perturbations. Asynchronous rollouts store synchronized observation histories, future body states, teacher actions, and task context to form the initial offline dataset $\mathcal D_0$. Complete rollout trajectories are split into training and evaluation sets before extracting fixed-length windows; normalization statistics are fitted only on the training split. Thus, no rollout contributes windows to both partitions. The teacher tracker is a training-time data source, not a component of PredActor's inference path.

\subsubsection{Teacher trajectory shooting and aggregation}

The aggregation update is given in Equation~\ref{eq:dagger_training}. The teacher continuation, not the student's future actions, supplies the target trajectory. The update adds supervised windows whose origins were visited during learner rollout; it does not change the diffusion objective or introduce reward optimization.

The horizon-teacher procedure in Figure~\ref{fig:method_training}(b) implements this training recipe and provides joint trajectory supervision for Equation~\ref{eq:dagger_training}. At round $j$, a frozen student advances the live simulator. A rollout origin $\xi$ contains the simulator state, student observation/action history, and motion reference phase needed to start an aligned teacher continuation. The tracking teacher is queried in a separate simulation branch with the same origin and reference. State and random-number-generator checks ensure that the query does not advance or alter the live student trajectory. Thus teacher actions label the branch; they are not interventions in the live rollout.

The window constructor $W_E$ retains the observed student-history prefix and appends the teacher continuation. Past actions are the student's executed actions paired with their pre-transition states; the action at the branch origin and subsequent actions come from the teacher. State targets follow the same temporal alignment. Task labels are aligned with the reference, and incomplete continuations are marked so that only complete training windows enter replay. Stored-window prefix lengths are configuration details and do not change the horizon-position convention $i=1,\ldots,h$. Each set of collected windows is used for continued diffusion reconstruction before the next collection round.

\subsubsection{Joint reconstruction objective}

Equation~\ref{eq:prdp_training} is used for both initial training and subsequent replay. Its expectation covers windows from $\mathcal D$, sampled noise, and diffusion levels. The action-difference term compares consecutive predicted and target actions, while the state reconstruction term supervises the internal future-state trajectory used for guidance. Iterative collection changes which windows enter replay without replacing this reconstruction objective.

\subsection{Task and Deployment Interface}
\label{app:sec:task_interface}
\label{app:deployment_details}

Task-specific providers inject raw condition keys on every step, and a composite provider distributes updates to its children. A condition composer then assembles the history consumed by the policy. This division allows observation-only and semantic/command-conditioned actors to share the same policy interface without hard-coding a particular input source.

The exported artifact contains a TorchScript backbone, diffusion schedule, normalizer statistics, actor configuration, and deployment configuration. The C++ runtime normalizes observations and conditions, runs the actor, unnormalizes the action trajectory, and forms the delay-compensated PD target by linearly interpolating the two clean action slots surrounding $u=(n_{\mathrm{obs}}-1)+\delta/\Delta t$. An asynchronous UDP receiver latches the latest external condition under a mutex, so condition transport is decoupled from the control loop. A Passive--Ready--Running state machine gates policy actions and falls back to Passive on fall or torque faults.

\subsection{Evaluated Deployment Profile}
\label{app:deployment_profile}

The evaluated G1 Orin profile is a batch-one FP32 conditioned PredActor policy with two-step deterministic DDIM, packed self- and cross-attention projections, cached command embeddings, a precomputed schedule, stacked-axis WBG, eager finite checks, and LibTorch inference mode.

For each level $t_i$, the reduced backbone predicts clean action and state trajectories once,
\begin{equation}
 (\hat a_i^0,\hat s_i^0)=F_\theta(a_i,s_i,t_i,c_{\mathrm{enc}}).
\end{equation}
When guidance is active, the same state prediction is adjusted by an analytic whole-body cost and reused in the DDIM update,
\begin{equation}
 \tilde s_i^0=\hat s_i^0-G(\hat s_i^0;u_{v_x},u_{v_y},u_{v_z}),\qquad
 (a_{i-1},s_{i-1})=\mathcal D(a_i,s_i,\hat a_i^0,\tilde s_i^0;t_i,t_{i-1}).
\end{equation}
The same clean state prediction is reused by guidance, and $\hat a_i^0$ is not recomputed at level $t_i$. The modified state enters $s_{i-1}$, so the next backbone evaluation can propagate the steering effect to its action prediction through state-to-action attention. The evaluated path therefore performs one backbone evaluation per diffusion step, or two in total. A duplicate-backbone variant executes four forwards and changes the guidance equation, so it is excluded from the exact acceleration ablation. The actor also performs encoding, guidance, scheduler updates, and boundary checks, so operation counts do not determine complete-loop speedups. Actor and callback timings, their acceptance thresholds, and the nonadditive component diagnostics are reported in Section~\ref{sec:orin_acceleration} and Appendix~\ref{app:orin_details}.

\section{Detailed Experimental Protocol}
\label{app:experiment_details}
\label{app:extended_experimental_design}

\subsection{Outcome Definitions and Missingness}
\label{app:metrics}

\textbf{Survival and recovery.} Architecture-study push survival is the
observed fraction of the 650-step horizon, so an early terminal contributes its observed duration rather than a binary zero; it is not the fraction of cells that finish. A terminal is triggered by root height below 0.55~m, absolute pelvis tilt at least 0.3~rad, or base angular speed at least 1~rad/s. Strict recovery requires 25 sustained post-force steps inside all three gates. In the architecture study, early failures remain in every denominator.

\textbf{Destination and text outcomes.} Destination arrival is first entry
within 0.6~m of the target. Final and minimum error use horizontal root-to-target distance. Text retrieval is raw top-1 MotionCLIP prompt retrieval over complete windows beginning at least 3~s after command activation. A trial may therefore contribute to survival but lack an eligible retrieval window. Generated references outside the tracker converter's accepted joint range are retained as non-evaluable rather than assigned failure or zero.

\textbf{Timing.} The timing endpoints in Figure~\ref{fig:orin-acceleration} are
$\mathrm{E}_0$ (actor execution with a fixed input tensor), $\mathrm{E}_1$ (receipt and conversion of a new robot state followed by actor execution), and $\mathrm{E}_2$ (the complete callback through production of the host action tensor). These boundaries are measured directly; their percentiles are not added. The 20~ms line is the 50~Hz target used for percentile and deadline-attainment reporting.

\subsection{Architecture Protocol and Coverage}
\label{app:architecture_protocol}

The six learned policies use the same 7,600-episode dataset, comprising 50 teacher rollouts for each of 152 G1-retargeted ACCAD motions. Training uses 50 epochs, a base learning rate of $10^{-4}$, cosine decay with 10,000 warm-up steps, batch size 256, data sampling rate 0.01, and seeds 92025, 92026, and 92027. Reported means and sample SDs are computed across the three checkpoint means. ARDY+SONIC is a single-model reference.

CLoC+FK and CLoC+FK w TextCFG denote the evaluated Diffuse-CLoC implementations with forward kinematics used to estimate the privileged-state inputs expected by the Diffuse-CLoC observation profile. The TextCFG variant additionally includes learned text conditioning and classifier-free guidance. The FK suffix describes only this observation interface. PredActor instead conditions on proprioceptive history, without FK-based reconstruction of privileged-state inputs.

\begin{center}
\begin{minipage}{\textwidth}
  \captionsetup{type=table}
  \centering
  \caption{Key Q1 policy settings. S+A denotes joint future-state and action prediction; A denotes action-only prediction. Text/CFG records the learned text-conditioning and classifier-free-guidance interface, whereas state CG records the predicted-state classifier-guidance interface used for destination steering.}
  \label{tab:q1-policy-settings}
  \small
  \setlength{\tabcolsep}{3.2pt}
  \resizebox{\linewidth}{!}{%
  \begin{tabular}{@{}lcccccc@{}}
    \toprule
    Policy & Output & Text/CFG & State CG & Observation/condition profile & Sampler & $n_{\mathrm{train}}$ \\
    \midrule
    DiffuseLoco w TextCFG & A & yes & no & \texttt{g1\_rlobs} & DDPM-20 & 3 \\
    CLoC+FK & S+A & no & yes & \texttt{g1\_standard} & DDPM-20 & 3 \\
    CLoC+FK w TextCFG & S+A & yes & yes & \texttt{g1\_standard} & DDPM-20 & 3 \\
    PredActor w/o TextCFG & S+A & no & yes & \texttt{g1\_prdp\_standard} & NS-DDIM-2 & 3 \\
    PredActor & S+A & yes & yes & \texttt{g1\_prdp\_standard\_cond} & NS-DDIM-2 & 3 \\
    PredActor w/o state & A & yes & no & \texttt{g1\_prdp\_standard\_cond} & NS-DDIM-2 & 3 \\
    ARDY+SONIC & reference $\rightarrow$ A & reference & no & \texttt{sonic\_release} & ARDY-H8 & 1 \\
    \bottomrule
  \end{tabular}}
\end{minipage}
\end{center}

PredActor w/o TextCFG and PredActor both predict a 20-step state--action trajectory with full cross-attention and two deterministic NS-DDIM steps. PredActor w/o TextCFG receives a 96-D proprioceptive condition through \texttt{g1\_prdp\_standard}; PredActor adds a 512-D text latent through \texttt{g1\_prdp\_standard\_cond}, giving a 608-D condition. Both use data sampling rate 0.01 during training. Their destination controllers use $v_x$ ranges of $[0,5]$ and $[0,8]$~m/s, respectively.

The matched state-prediction ablation pairs separately retrained joint state--action and action-only checkpoints by seed under a primary no-CG contract. The matched joint state--action arm is distinct from the configured PredActor row in Table~\ref{tab:q1-architecture-comparison}; the action-only row in that table is the other arm. Table~\ref{tab:q1-state-ablation} reports the paired comparison.

\begin{center}
\begin{minipage}{\textwidth}
  \captionsetup{type=table}
  \centering
  \caption{Matched state-prediction ablation under the primary no-CG contract. Both arms are separately retrained for this comparison. Arm entries are mean $\pm$ sample SD across three training checkpoints. Effects are matched joint state--action minus action-only, with two-sided 95\% $t$ intervals across the three paired seed differences.}
  \label{tab:q1-state-ablation}
  \small
  \setlength{\tabcolsep}{5pt}
  \begin{tabular}{@{}lccc@{}}
    \toprule
    Metric & Matched PredActor & PredActor w/o state & Paired effect [95\% CI] \\
    \midrule
    Push survival $\uparrow$ & $0.535\pm0.009$ & $0.487\pm0.033$ & $+0.048\;[-0.024,\,0.120]$ \\
    Push recovery $\uparrow$ & $0.043\pm0.040$ & $0.193\pm0.110$ & $-0.150\;[-0.494,\,0.194]$ \\
    Text survival $\uparrow$ & $0.866\pm0.011$ & $0.778\pm0.145$ & $+0.088\;[-0.246,\,0.421]$ \\
    Text retrieval $\uparrow$ & $0.538\pm0.087$ & $0.397\pm0.023$ & $+0.142\;[-0.019,\,0.302]$ \\
    Jerk RMS $\downarrow$ & $15.684\pm2.368$ & $12.448\pm0.675$ & $+3.236\;[-2.200,\,8.671]$ \\
    Latency at $B=5$ (ms) $\downarrow$ & $1.335\pm0.039$ & $1.050\pm0.004$ & $+0.285\;[0.199,\,0.372]$ \\
    \bottomrule
  \end{tabular}
  \vspace{0.35em}

  \fontsize{7}{8}\selectfont
  Both arms use the same data, observation and condition tensors, 20-step action horizon, optimizer and schedule, two-step sampler, runtime, and seeds. Trainable parameter counts differ by 0.0161\%. Destination metrics are unavailable because online state CG is disabled. Retrieval uses 113/135 eligible windows for matched PredActor and 98/135 for the action-only arm.
\end{minipage}
\end{center}

The architecture study uses one calibrated MuJoCo plant with whole-body guidance disabled outside the destination task. Push evaluation applies 100, 200, 300, 400, or 500~N horizontally for 0.5~s during walking. Each checkpoint uses the same 20 paired direction/seed draws at every force, giving 100 requested cells; the table therefore contains 300 cells for a three-checkpoint row and 100 for a single-model row. Destination following uses three target draws for each of five evaluation seeds and runs for at most 20~s. Text evaluation crosses nine prompts with the five evaluation seeds. Actor latency is synchronized on one exclusive RTX~4090, measured after warm-up, and reported per action from a fixed policy input repeated to batch five. Basic walking jerk uses the paired pre-push nominal segment: 20 cells per checkpoint for the original architecture-suite rows and 25 cells per checkpoint for PredActor w/o state from its matched state-ablation suite, yielding 60/60 and 75/75, respectively. Table~\ref{tab:architecture-coverage} records evaluated and requested cells after combining training replicas without changing the outer uncertainty unit.

\begin{center}
\begin{minipage}{\textwidth}
  \captionsetup{type=table}
  \centering
  \caption{Architecture-study coverage. Counts are evaluated/requested cells.}
  \label{tab:architecture-coverage}
  \small
  \setlength{\tabcolsep}{3.5pt}
  \begin{tabular}{@{}lcccc@{}}
    \toprule
    Policy & Push & Destination & Text survival & Text retrieval \\
    \midrule
    DiffuseLoco w TextCFG & 300/300 & 0/45 & 135/135 & 126/135 \\
    CLoC+FK & 300/300 & 45/45 & 0/135 & 0/135 \\
    CLoC+FK w TextCFG & 300/300 & 45/45 & 135/135 & 103/135 \\
    PredActor w/o TextCFG & 300/300 & 45/45 & 0/135 & 0/135 \\
    PredActor & 300/300 & 45/45 & 135/135 & 119/135 \\
    PredActor w/o state & 300/300 & 0/45 & 135/135 & 98/135 \\
    ARDY+SONIC & 100/100 & 2/15 & 23/45 & 23/45 \\
    \bottomrule
  \end{tabular}
\end{minipage}
\end{center}

\begin{figure}[ht]
  \centering
  \includegraphics[width=0.82\textwidth]{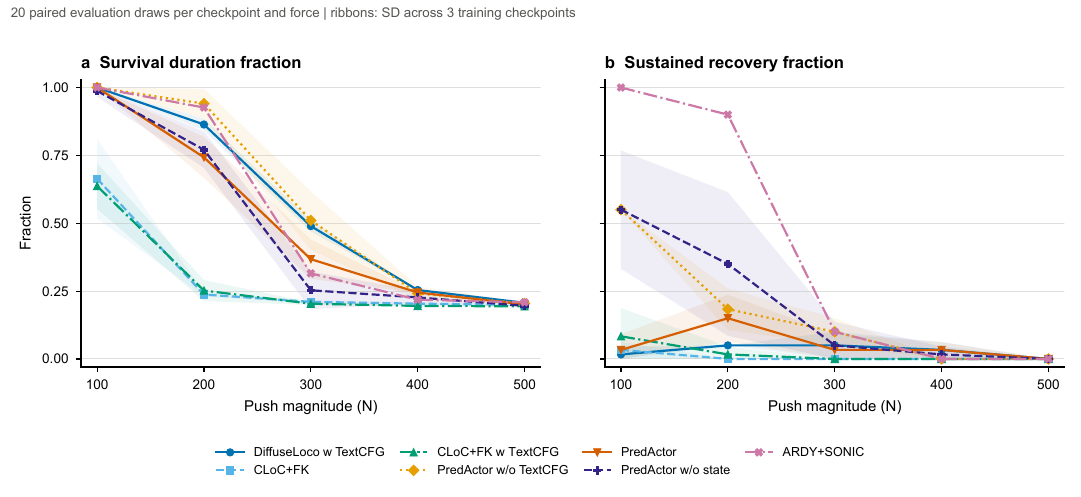}
  \caption{\textbf{Architecture survival and recovery over all tested push
  magnitudes.} Each checkpoint contributes 20 paired evaluation draws at each
  force. Points are means across checkpoint means and ribbons show sample SD
  across the three training checkpoints. ARDY+SONIC has one checkpoint and no
  ribbon. Table~\ref{tab:q1-architecture-comparison} averages all five forces.}
  \label{fig:architecture-push-supplement}
\end{figure}

Figure~\ref{fig:architecture-push-supplement} resolves the aggregate push results by force magnitude.

\FloatBarrier
\subsection{Deployment Demonstration Scope}
\label{app:deployment_demo_protocol}

The demonstration suite spans simulation and hardware. Joystick steering and semantic interpolation are evaluated in simulation (Figure~\ref{fig:hoffman_demos}(a,c)); external interference and staged text commands are demonstrated on the physical Unitree G1 (Figure~\ref{fig:hoffman_demos}(b,d)). Figure~\ref{fig:orin-acceleration}(g) provides additional onboard frames from the G1--Jetson Orin NX stack. The deployed actor uses the observation, condition, export, and safety interfaces described in Appendix~\ref{app:deployment_details}.

\begin{figure}[t]
  \centering
  \includegraphics[width=\linewidth]{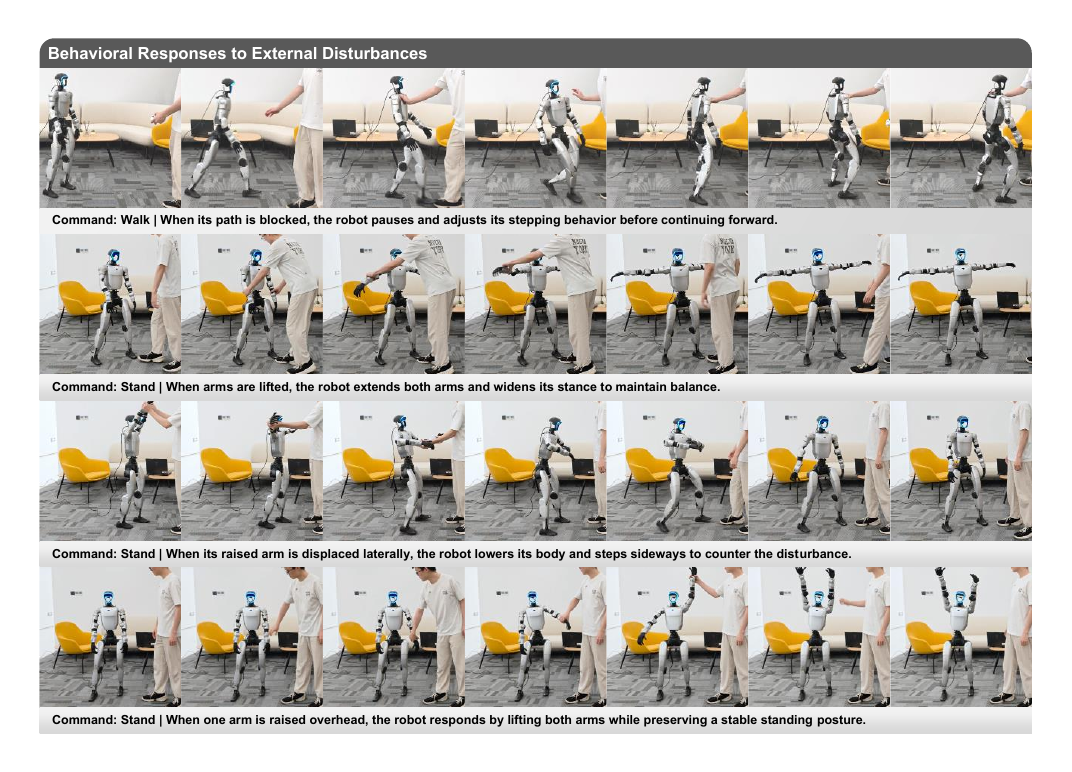}
  \caption{Qualitative behavioral responses of the physical Unitree G1 under external interaction. Each sequence identifies the active command and the observed response: the walking policy adjusts its stepping when obstructed, while the standing policy coordinates arm extension, stance widening, body lowering, and lateral stepping in response to upper-body or arm displacement. These selected successful sequences illustrate the observed response modes.}
  \label{fig:behavior-response-appendix}
\end{figure}

The selected physical sequences show successful integrated execution, while Table~\ref{tab:real-robot-quant} and Appendix~\ref{app:real_robot_trial_sequences} report all five outcomes per setting.

\section{Complementary Results and Limitations}
\label{app:complementary_results}

\subsection{Orin Sampler and Implementation Ablations}
\label{app:orin_details}
\label{app:q2_admissibility}
\label{app:q2_checkpoint_artifacts}

The sampler sweep in Table~\ref{tab:orin-sampler} uses $\mathrm{P}_0$, the original TorchScript actor before the acceleration changes evaluated here. Rolling DDPM has 20 schedule levels but executes 15 updates in the deployed rolling window. With legacy WBG it evaluates the backbone twice per update. The two-step WBG path executes two backbone forwards in total. Figure~\ref{fig:orin-acceleration}(a) reports the timing sweep, and Figure~\ref{fig:orin-acceleration}(d) separately measures push survival across the tested sampler budgets; together they identify DDIM-2 as the best tested trade-off between inference speed and disturbance survival.

\begin{center}
\begin{minipage}{\textwidth}
  \captionsetup{type=table}
  \centering
  \caption{$\mathrm{P}_0$ actor timing on Jetson Orin. Values are actor CUDA
  p50 in milliseconds, each the median of three block medians.}
  \label{tab:orin-sampler}
  \small
  \setlength{\tabcolsep}{4pt}
  \begin{tabular}{@{}lrrr@{}}
    \toprule
    Sampler & Steps & WBG off & WBG on \\
    \midrule
    NS-DDIM & 1 & 9.725 & 12.221 \\
    NS-DDIM & 2 & 16.148 & 21.846 \\
    NS-DDIM & 3 & 22.915 & 29.946 \\
    NS-DDIM & 5 & 36.001 & 46.948 \\
    NS-DDIM & 10 & 68.091 & 92.011 \\
    Rolling DDPM & 15 & 121.229 & 235.216 \\
    \bottomrule
  \end{tabular}
\end{minipage}
\end{center}

The exact factorial study crosses four attention-projection layouts with condition-embedding reuse, DDIM schedule precomputation, and sequential versus stacked WBG axes. All 32 valid combinations use the production one-backbone-per-denoising-step guidance update (two backbone forwards total for NS-DDIM2). Across matched valid cells, all-QKV packing saves 0.775~ms with a range of [0.256, 0.799]~ms across the three block-level effects; condition reuse saves 0.190~ms, with a range of [0.151, 0.347]~ms; precomputation saves 1.672 [1.477, 1.698]~ms, and stacked WBG saves 2.845 [2.627, 2.868]~ms. These conditional effects depend on the other settings and must not be summed. The cumulative path in Table~\ref{tab:orin-implementation-ablation} reports individual matched composition differences rather than these matrix-wide conditional effects. Cross-attention-only packing is not consistently faster. An implementation that duplicated backbone execution changes the guidance equation and is excluded from the exact ablation.

The optimization path also crosses two controlled boundaries. The original-export sampler endpoint $P_0=21.846$~ms and the matched structural-ablation baseline of 21.461~ms belong to separate suites. Within the structural-ablation suite, stacked WBG reduces this matched baseline to $P_1=18.315$~ms; $P_0$ is not differenced against $P_1$. The factorially selected exact configuration $P_2$ measures 15.930~ms. The same configuration measures 16.234~ms as $P_3$ in the runtime suite; the difference reflects the measurement suite rather than an implementation change. Enabling LibTorch inference mode at $P_4$ reduces the median by 3.058~ms to 13.175~ms.

Table~\ref{tab:orin-comprehensive-ablation} collects all 32 valid structural compositions as 16 sequential/stacked WBG pairs, all eight runtime controls, and matched component effects. Frozen actor and live callback timings remain separate; detailed per-boundary records are retained in the source data.

\clearpage
\begin{center}
\begin{minipage}{\textwidth}
\captionsetup{type=table}
\caption{\textbf{Comprehensive exact Orin implementation ablation.} NS-DDIM2, B=1 FP32, WBG on, two backbone forwards. Times are milliseconds.}
\label{tab:orin-comprehensive-ablation}
\centering\small
\setlength{\tabcolsep}{3pt}\renewcommand{\arraystretch}{1.0}
\textbf{A. All 32 structural compositions (16 paired rows).}\par\vspace{2pt}
\begin{tabular*}{\linewidth}{@{}l@{\extracolsep{\fill}}ccrrrrr@{}}
\toprule
QKV & C & P & Seq. p50 & Seq. p95 & Stk. p50 & Stk. p95 & Saved$_{\rm WBG}$ \\
\midrule
none & 0 & 0 & 21.461 & 22.036 & 18.315 & 19.677 & 3.147 \\
self & 0 & 0 & 21.233 & 22.260 & 17.972 & 18.376 & 3.261 \\
cross & 0 & 0 & 20.669 & 22.016 & 18.583 & 18.885 & 2.086 \\
all & 0 & 0 & 20.522 & 20.812 & 17.751 & 18.873 & 2.771 \\
none & 0 & 1 & 19.082 & 20.631 & 17.442 & 18.347 & 1.640 \\
self & 0 & 1 & 19.315 & 20.292 & 16.232 & 17.116 & 3.083 \\
cross & 0 & 1 & 19.674 & 20.656 & 17.180 & 17.580 & 2.494 \\
all & 0 & 1 & 18.685 & 19.374 & 16.331 & 17.372 & 2.354 \\
none & 1 & 0 & 20.985 & 21.948 & 18.188 & 19.359 & 2.798 \\
self & 1 & 0 & 20.384 & 21.196 & 17.789 & 19.430 & 2.595 \\
cross & 1 & 0 & 20.717 & 21.823 & 17.888 & 18.414 & 2.829 \\
all & 1 & 0 & 21.112 & 22.008 & 17.423 & 18.262 & 3.688 \\
none & 1 & 1 & 19.184 & 19.492 & 16.631 & 17.572 & 2.553 \\
self & 1 & 1 & 19.258 & 20.695 & 16.104 & 16.825 & 3.154 \\
cross & 1 & 1 & 19.177 & 19.492 & 16.387 & 16.720 & 2.790 \\
all & 1 & 1 & 18.803 & 19.467 & \textbf{15.930} & \textbf{16.728} & 2.873 \\
\bottomrule\end{tabular*}
\par\vspace{2pt}\begin{minipage}{\linewidth}\footnotesize C: condition cache; P: DDIM precomputation (0/1 = off/on). Seq./Stk.: sequential/stacked WBG. Saved$_{\rm WBG}$ = Seq. p50 $-$ Stk. p50 in the same cell; inference off, eager checks.\end{minipage}
\par\vspace{6pt}
\textbf{B. All eight runtime configurations (separate suite).}\par\vspace{2pt}
\begin{tabular*}{\linewidth}{@{}l@{\extracolsep{\fill}}clrrrrr@{}}
\toprule
Structure & I & Checks & Actor p50 & p95 & Saved & Host p50 & $>20$ ms \\
\midrule
base & 0 & eager & 18.266 & 19.104 & 0.000 & 21.042 & 600/600 \\
base & 0 & defer & 18.301 & 19.286 & -0.035 & 20.943 & 600/600 \\
base & 1 & eager & 15.401 & 15.651 & 2.865 & 19.451 & 49/600 \\
base & 1 & defer & 14.924 & 15.201 & 3.342 & 18.873 & 61/600 \\
all & 0 & eager & 16.234 & 16.912 & 2.032 & 18.855 & 60/600 \\
all & 0 & defer & 15.374 & 15.932 & 2.892 & 18.446 & 90/600 \\
\textbf{all} & \textbf{1} & \textbf{eager} & \textbf{13.175} & \textbf{13.400} & \textbf{5.090} & \textbf{16.790} & \textbf{7/600} \\
all & 1 & defer & 13.183 & 13.609 & 5.083 & 16.836 & 13/600 \\
\bottomrule\end{tabular*}
\par\vspace{2pt}\begin{minipage}{\linewidth}\footnotesize Base/all: QKV packing, cache and precompute all off/on; WBG stacked. I: inference mode. Saved compares frozen actor p50 to the first row (18.266 ms). Host = live callback-to-host; last column counts host deadline misses. Deferred checks change failure-detection semantics.\end{minipage}
\par\vspace{6pt}
\textbf{C. Conditional frozen-actor savings, not additive.}\par\vspace{2pt}
\begin{tabular*}{\linewidth}{@{}l@{\extracolsep{\fill}}lrrr@{}}
\toprule
Component & Matched change & Pairs/block & Saved & Block-effect range \\
\midrule
QKV packing & none $\to$ self & 8 & 0.221 & [0.003, 0.555] \\
QKV packing & none $\to$ cross & 8 & 0.172 & [-0.100, 0.363] \\
QKV packing & none $\to$ all & 8 & 0.775 & [0.256, 0.799] \\
Condition cache & off $\to$ on & 16 & 0.190 & [0.151, 0.347] \\
DDIM precompute & off $\to$ on & 16 & 1.672 & [1.477, 1.698] \\
WBG axes & sequential $\to$ stacked & 16 & 2.845 & [2.627, 2.868] \\
Inference mode & off $\to$ on (selected structure) & 1 & 3.166 & [2.812, 3.492] \\
\bottomrule\end{tabular*}
\par\vspace{3pt}\begin{minipage}{\linewidth}\footnotesize Each primary cell uses three rotated blocks of 200 retained calls; p50/p95 are medians of block quantiles. Panel C gives median matched block effects and their min--max range, not confidence intervals. Its inference-mode paired effect (3.166 ms) differs from the 3.058 ms difference of endpoint medians shown in Figure~\ref{fig:orin-acceleration}(e) and Table~\ref{tab:orin-implementation-ablation}. Bold marks the selected structure/runtime. The E1 new-state-plus-actor endpoint has p50/p95 of 13.843/16.221 ms; callback p95 is 19.383 ms, with 7/600 calls over 20 ms. Sampler controls remain in Table~\ref{tab:orin-sampler}; the complete per-boundary source data retain block ranges, actor misses and live structural timings omitted here. The 32 duplicate-backbone cells are excluded as an execution bug; no cross-suite savings are inferred.\end{minipage}
\end{minipage}
\end{center}

\clearpage

\subsection{Timing Decomposition}
\label{app:timing_decomposition}

Table~\ref{tab:orin-components} records the median component diagnostics most relevant to the deployment boundary. The two backbone and WBG spans are inclusive traced CUDA regions. Their medians cannot be added to each other or to host spans. Tracing inflates the final actor span to 40.090~ms, whereas an independent unprofiled fixed-input pass measures 12.198~ms; neither replaces the primary 13.175~ms $\mathrm{E}_0$ endpoint.

\begin{center}
\begin{minipage}{\textwidth}
  \captionsetup{type=table}
  \centering
  \caption{Diagnostic component p50 timing in milliseconds.}
  \label{tab:orin-components}
  \small
  \setlength{\tabcolsep}{4pt}
  \begin{tabular}{@{}lrrl@{}}
    \toprule
    Component & $\mathrm{P}_0$ & $\mathrm{P}_4$ & Boundary \\
    \midrule
    State fetch/copy & 0.002 & 0.001 & host \\
    Pre-FK and FK & 0.757 & 0.779 & host \\
    Observation construction & 0.919 & 0.997 & host \\
    Condition packing & 0.083 & 0.082 & host \\
    Action selection/copy & 0.234 & 0.238 & host \\
    Backbone forward 1 & 18.053 & 13.056 & traced CUDA \\
    Backbone forward 2 & 18.253 & 13.852 & traced CUDA \\
    WBG gradient 1 & 3.095 & 2.562 & traced CUDA \\
    WBG gradient 2 & 3.030 & 2.648 & traced CUDA \\
    Actor-loop residual & 8.054 & 2.474 & traced CUDA \\
    \bottomrule
  \end{tabular}
\end{minipage}
\end{center}

\subsection{Capability, Coverage, and Interpretation Limits}
\label{app:experiment_limitations}

\textbf{Architecture study.} Six policies each use three independently trained
checkpoints with seeds 92025/92026/92027. Means and sample SDs are taken across
checkpoint means. Each checkpoint contributes 20 paired evaluation draws at
each of five push forces, 15 destination trials, 45 text trials, and five timing
runs where the policy supports the corresponding task. ARDY+SONIC uses one
checkpoint; 2/15 destination references and 23/45 text references pass its
joint-range converter. MotionCLIP retrieval measures agreement between the
executed motion and the requested text.

\textbf{Timing study.} Timing applies to one batch-one FP32 artifact on one
Jetson Orin. The benchmark measures the offline callback-to-host-action boundary from received robot-state messages; all actuation counters remain zero. The complete-callback p95 is below 20~ms, and 593/600 calls meet the target; the seven observed overruns define the remaining tail-latency margin. Physical closed-loop behavior is documented by the deployment demonstrations. Graph bytes, thermal state, runtime version, and implementation settings are part of the measured system; a different export or device requires requalification.

\textbf{Real-robot demonstrations and scope.} Figure~\ref{fig:hoffman_demos}(d) documents staged text-command execution, whereas Figure~\ref{fig:hoffman_demos}(b) and Figure~\ref{fig:behavior-response-appendix} show selected responses to external interaction. Figure~\ref{fig:orin-acceleration}(g) separately provides qualitative onboard frames from the G1--Jetson Orin NX stack. Table~\ref{tab:real-robot-quant} gives descriptive success counts from five trials per transition or push setting, and Figures~\ref{fig:trial-seq-stand-squat}--\ref{fig:trial-seq-walk-push} show all recorded outcomes. Success requires completing the requested transition or remaining upright after the push; falls, incomplete behaviors, and safety stops are failures.

\textbf{Future scope.} The current guidance interface is calibrated over the tested objective range; stronger guidance can move denoising outside the policy's training distribution. Text control is limited to learned semantic categories because the conditioning model and the coverage of the training data and motion library do not yet support reliable interpretation of complex or compositional commands. The training library also does not cover sustained multi-contact behaviors such as crawling or climbing. Future work will investigate constrained or adaptive guidance, broader language--motion data, and training objectives for richer contact transitions.

\subsection{Real-Robot Trial Sequences}
\label{app:real_robot_trial_sequences}

\begin{figure*}[p]
  \centering
  \includegraphics[width=0.92\textwidth]{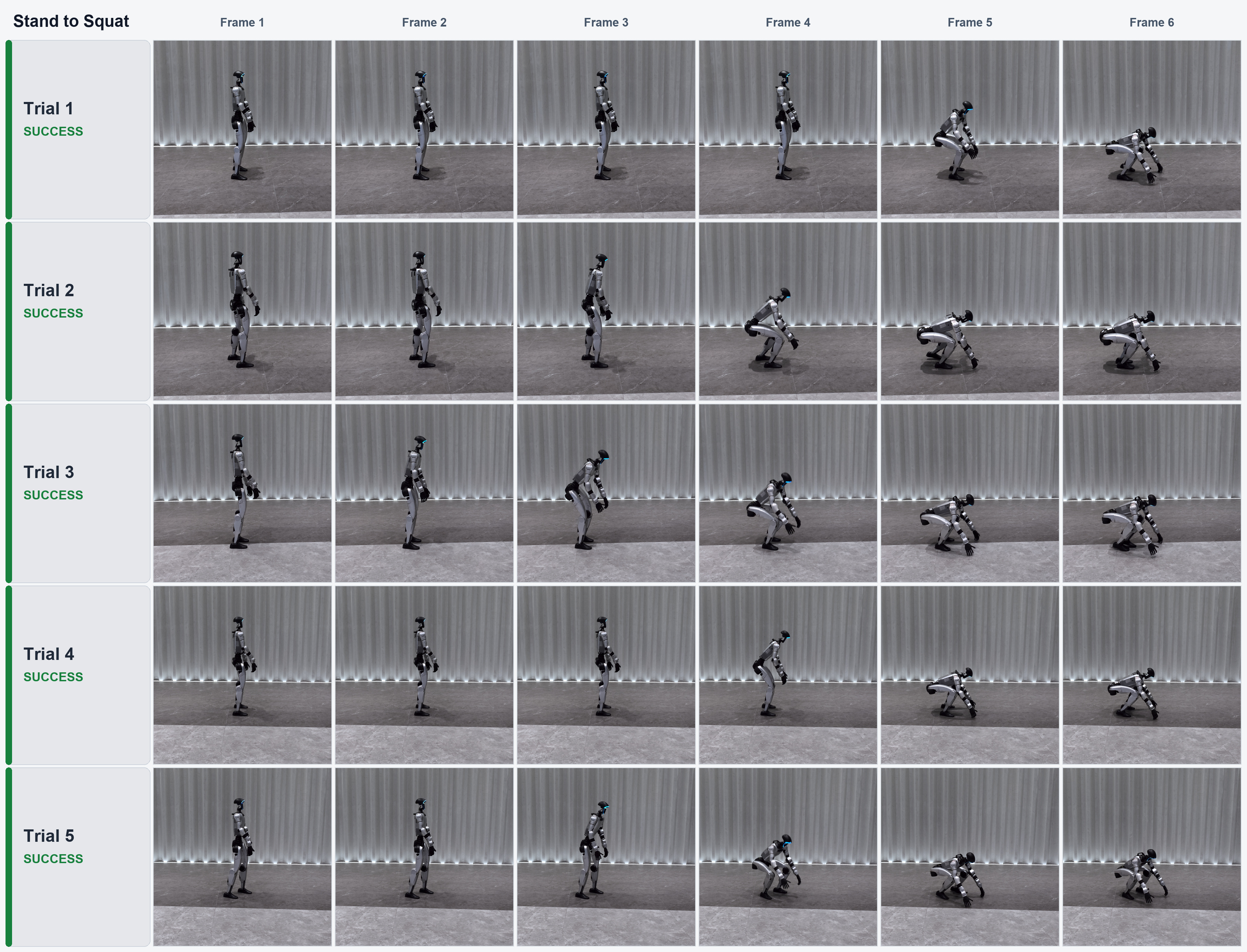}
  \caption{Stand-to-squat trials. Each row is one trial; columns progress from left to right.}
  \label{fig:trial-seq-stand-squat}
\end{figure*}

\begin{figure*}[p]
  \centering
  \includegraphics[width=0.92\textwidth]{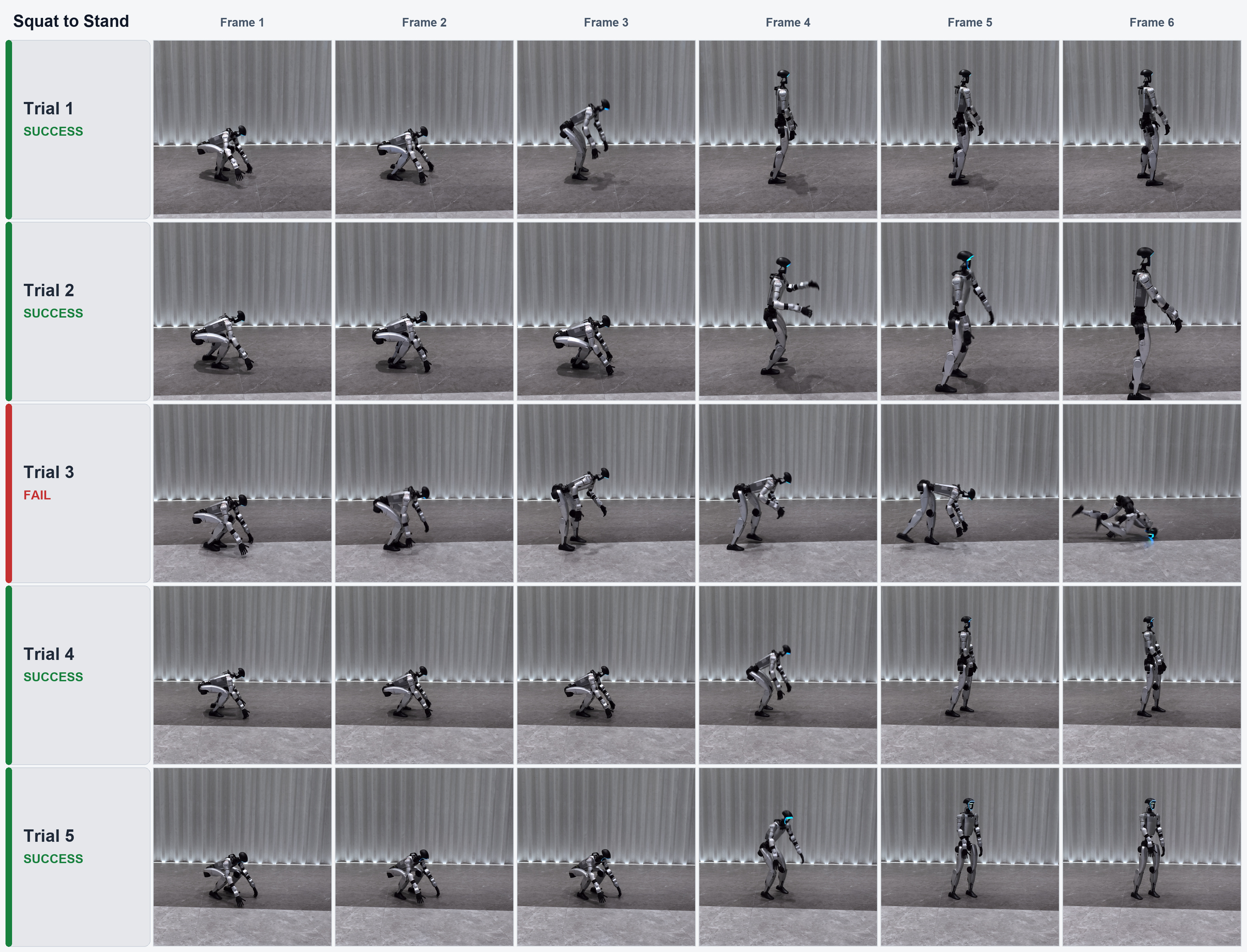}
  \caption{Squat-to-stand trials, including the failed third trial.}
  \label{fig:trial-seq-squat-stand}
\end{figure*}

\begin{figure*}[p]
  \centering
  \includegraphics[width=0.92\textwidth]{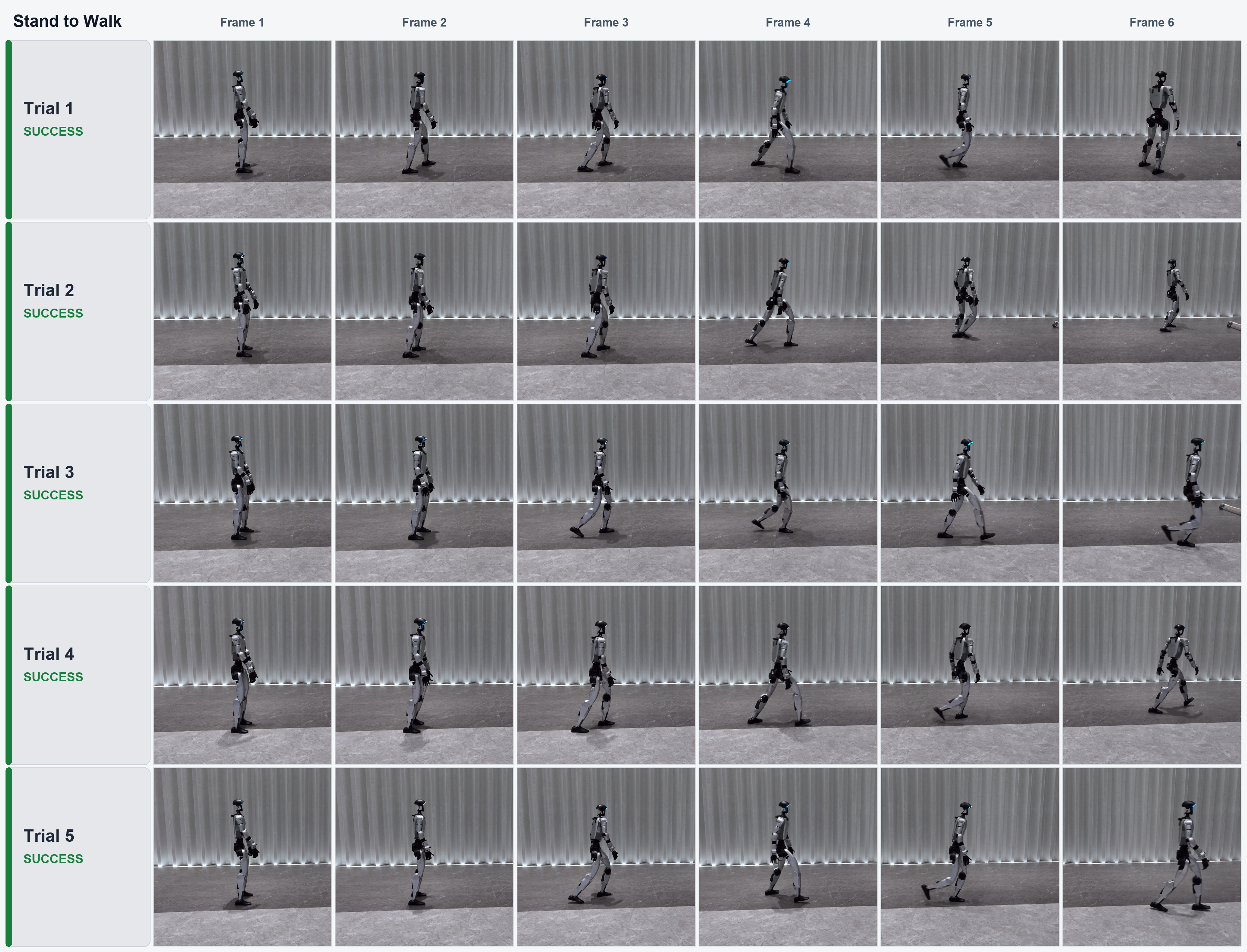}
  \caption{Stand-to-walk trials. The first column shows the stationary initial pose; the remaining columns show walking.}
  \label{fig:trial-seq-stand-walk}
\end{figure*}

\begin{figure*}[p]
  \centering
  \includegraphics[width=0.92\textwidth]{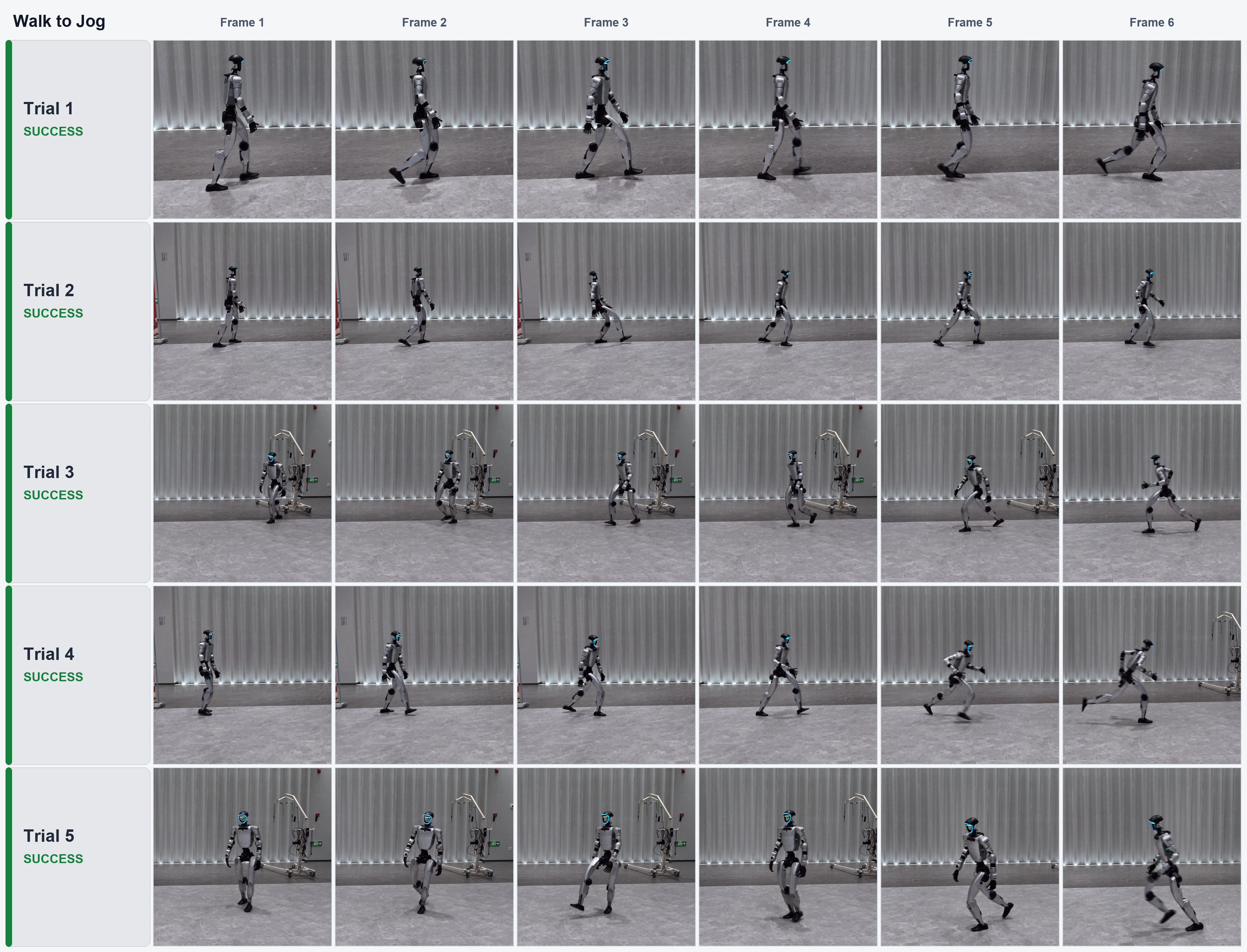}
  \caption{Walk-to-jog trials. Columns 1--3 show walking and the transition; columns 4--6 show the brief jogging phase.}
  \label{fig:trial-seq-walk-jog}
\end{figure*}

\begin{figure*}[p]
  \centering
  \includegraphics[width=0.92\textwidth]{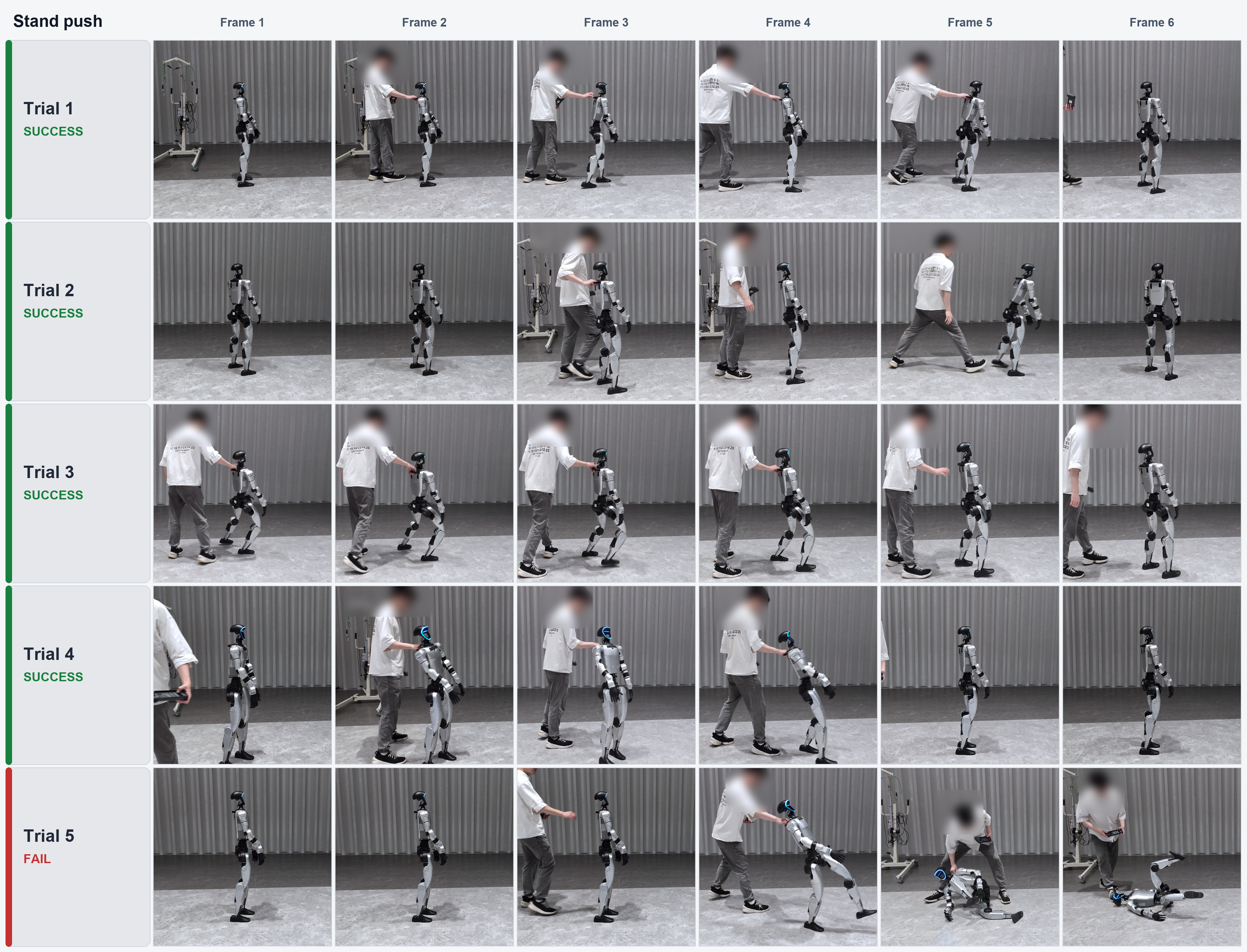}
  \caption{Standing push-response trials, including the failed fifth trial. Successful rows end after stable recovery; the failed row retains the fall outcome.}
  \label{fig:trial-seq-stand-push}
\end{figure*}

\begin{figure*}[p]
  \centering
  \includegraphics[width=0.92\textwidth]{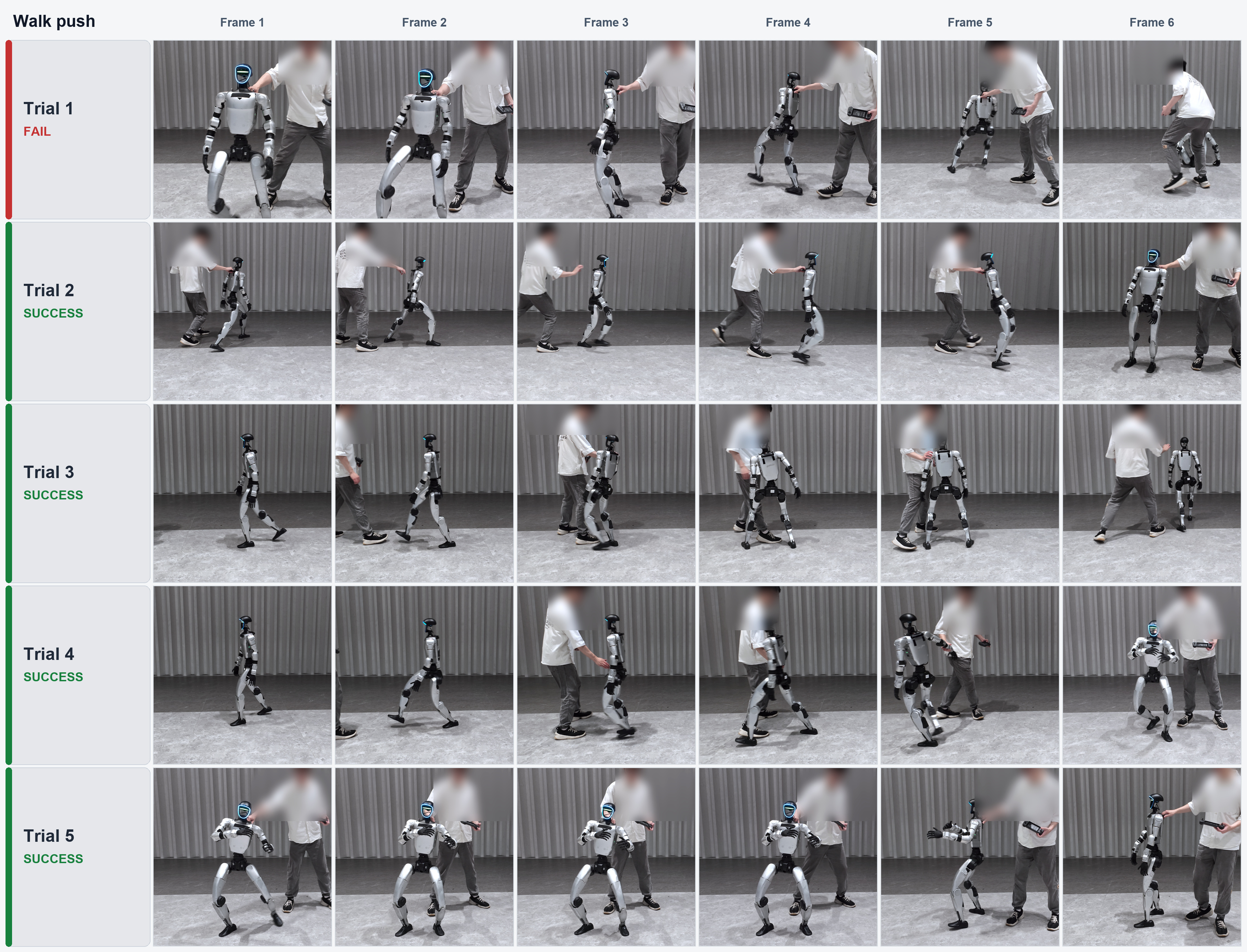}
  \caption{Walking push-response trials, including the failed first trial. Successful rows show walking, perturbation, and a final frame after stable walking resumes; the failed row retains the fall outcome.}
  \label{fig:trial-seq-walk-push}
\end{figure*}

Figures~\ref{fig:trial-seq-stand-squat}--\ref{fig:trial-seq-walk-push}
show six temporally ordered frames from each recorded trial for all six settings
reported in Table~\ref{tab:real-robot-quant}. The left rail reports the trial
index and outcome under the same success rule.

\FloatBarrier

\end{document}